\documentclass{article}

\PassOptionsToPackage{numbers, sort}{natbib}

\usepackage[main, final]{neurips_2026}

\usepackage[utf8]{inputenc} %
\usepackage[T1]{fontenc}    %
\usepackage{hyperref}       %
\usepackage{url}            %
\usepackage{booktabs}       %
\usepackage{amsfonts}       %
\usepackage{nicefrac}       %
\usepackage{microtype}      %
\usepackage{xcolor}         %
\usepackage{graphicx}
\usepackage{amsmath}
\usepackage{xspace}
\usepackage{wrapfig}
\usepackage{subcaption}

\newcommand\blfootnote[1]{%
  \begingroup
  \begin{NoHyper}%
  \renewcommand\thefootnote{}\footnote{\color{black}#1}%
  \addtocounter{footnote}{-1}%
  \end{NoHyper}%
  \endgroup
}

\DeclareMathOperator{\csim}{sim}

\title{Do Sparse Autoencoders Learn Meaningful \\ Concept Hierarchies?}

\author{Nils Grandien$^{1, \ast}$
\and
\textbf{David Steinmann}$^{1,2, \ast}$
\and
\textbf{Felix Friedrich}$^{3}$
\and 
\textbf{Kristian Kersting}$^{1,2,4}$
\and
$^{1}$Computer Science Department, TU Darmstadt 
$^{2}$Hessian Center for AI (hessian.AI) \\
$^{3}$Black Forest Labs
$^{4}$German Research Center for AI (DFKI)
}

\begin{document}

\maketitle

\begin{abstract}
Sparse autoencoders (SAEs) have become an important tool for unsupervised concept discovery in large models. To make the resulting feature spaces more interpretable and manageable, recent approaches have begun imposing hierarchical structure, either explicitly or as an implicit effect of training constraints, yet rigorous comparison remains difficult. There are no agreed-upon requirements for what a meaningful feature hierarchy should satisfy, and evaluation has largely relied on qualitative illustrations with fragmented quantitative protocols. To address this, we derive a set of key requirements for generalization/specialization hierarchies in unsupervised concept discovery, drawing on semantic net and taxonomy research alongside recent SAE work, and use them to derive a concrete evaluation protocol. Applying this protocol to current SAE approaches trained on visual data, we find that while feature spaces generally provide a basis for sensible hierarchies, establishing good hierarchical structure remains challenging. 
In particular, feature absorption, both in its well-known hard form and in a continuous, soft form, systematically compromises hierarchy quality, pointing to a fundamental tension that future approaches will need to navigate.\blfootnote{\hspace{-0.15cm}$^{\ast}$These authors share equal contribution.}\footnote{The code for this work is available at: \url{https://github.com/nlsgrndn/eval-HSAEs}}

\end{abstract}

\section{Introduction}
A central challenge in deep learning is balancing model expressiveness with interpretability and user control.
Concept-based representations have emerged as a structured middle ground, supporting inherent explanation \citep{koh2020concept, kim2018interpretability, poeta2023concept} and direct model control \citep{liang2024controllable, teso2023leveraging}. 
Unsupervised concept discovery approaches, such as sparse autoencoders (SAEs), recover such representations without predefined labels \citep{fel2023holistic, rao2024discover, stammer2024neural, bricken2023monosemanticity, cunningham2023sparse, pach2025sparse}.

\begin{figure}[t]
    \centering
    \includegraphics[width=0.95\linewidth]{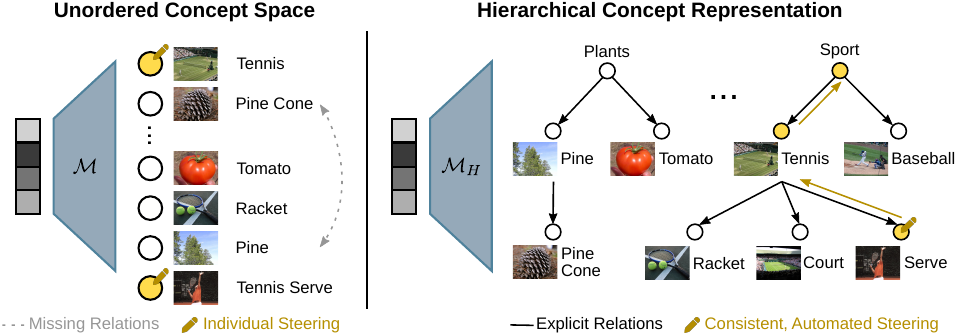}
    \caption{\textbf{Hierarchical concept representations explicitly model relations between concepts and allow for consistent model steering.} Typical unsupervised concept discovery approaches (left) provide large and unordered concept spaces. These lack clear connections between concepts, making interpretation difficult and cumbersome. Similarly, model steering requires identifying and updating many concepts at once. Hierarchical concept representations (right) instead explicitly provide a hierarchy graph for interpretation as well as consistent and automated model steering.}
    \label{fig:intro_figure}
\end{figure}

As these representations grow larger and more general, however, navigating and interacting with them becomes increasingly difficult \citep{poursabzi2021manipulating}, a limitation that stems in part from their lack of meaningful structure (cf.\ \autoref{fig:intro_figure}). Humans naturally organize concepts hierarchically \citep{miller1956magical, liu2004conceptnet}, increasing the interest in hierarchies within this context for their clear semantics and natural abstraction capabilities \citep{pittino2023hierarchical, sun2024eliminating, bussmann2025learning}. 
A number of methods now aim to learn such hierarchical structures within SAE representation spaces \citep{zaigrajew2025interpreting, costa2025flat, pach2025sparse, arad2025saes, li2025unlocking, bussmann2025learning}. 
Yet each relies on its own criteria and evaluation protocol, making meaningful comparison difficult and leaving the field without shared objectives for what a hierarchy in this setting should achieve.

In this work, we derive a set of general requirements for \textit{generalization/specialization} hierarchies in the context of unsupervised concept learning, drawing on established ideas from taxonomy evaluation \citep{tartir2010ontological} and semantic nets \citep{brachman1983and} alongside recently introduced metrics from the SAE literature \citep{luo2026atoms}. 
Importantly, we combine static, structural requirements, such as \textit{hierarchical abstractness}, with dynamic, co-activation-based requirements, such as \textit{frequent refinement}, the latter being particularly important in the context of unsupervised concept discovery. From these requirements, we derive a concrete set of metrics to assess the hierarchical capabilities of SAE-based concept extractors, which we use to evaluate current approaches that impose hierarchical structure either explicitly or implicitly during training. In doing so, we identify the current state of the field and concrete guidelines for developing future hierarchical concept extractors.

In summary, our contributions are (1) establishing a general set of requirements for \textit{generalization/specialization} hierarchies in unsupervised concept learning, (2) deriving concrete metrics to evaluate concept extractors against these requirements, (3) evaluating current SAE approaches trained on visual data on the quality of their hierarchical structure, and (4) identifying concrete guidelines for the future development of hierarchical concept extractors.

This work is structured as follows: We discuss related work in \autoref{sec:related_work}, before introducing our requirements and corresponding metrics in \autoref{sec:method}. We provide an experimental analysis of previous SAE approaches in \autoref{sec:experiments}, discuss main implications in \autoref{sec:discussion}, before concluding the paper in \autoref{sec:conclusion}.

\section{Related Work} \label{sec:related_work}

\paragraph{Concept Hierarchies.}
Concept hierarchies can be understood as a specific type of (directed, acyclic) knowledge graph \citep{abu2021domain} in which relations carry a unified semantics \citep{brachman1983and}. Common instantiations are \textit{is-a} hierarchies \citep{brachman1983and} and part-whole hierarchies \citep{keet2008representing}, both of which impose a clear generalization/specialization ordering on concepts. A long line of work has studied the automated construction of such hierarchies, also known as taxonomies, from text, ranging from lexico-syntactic patterns \citep{hearst1992automatic} to recent LLM-based approaches \citep{zeng2024chain, vu2025automated}, and has developed corresponding evaluation frameworks. These range from qualitative criteria such as conciseness and comprehensiveness \citep{nickerson2013method, unterkalmsteiner2023compendium} to quantitative supervised benchmarks \citep{bordea2016semeval}, and unsupervised measures such as information richness \citep{tartir2010ontological} or consistency between taxonomic distance and embedding similarity \citep{wullschleger2025no}. 
In unsupervised settings, concept hierarchies are related to hierarchical clustering approaches, with metrics such as cophenetic correlation \citep{farris1969cophenetic} or the cost metric by \citet{dasgupta2016cost}. 
For unsupervised representation spaces, 
\citet{park2025geometry} investigate the representation of hierarchical concepts directly and provide a perspective on the geometry of hierarchical concept representations.
Our work builds on this, but shifts the focus to hierarchies in the context of unsupervised concept discovery independent of representation geometry. This setting introduces an additional dimension absent from classical taxonomy evaluation: beyond static, structural properties, the hierarchy is instantiated dynamically by a concept encoder, raising the question of how concepts co-activate and refine one another on concrete inputs.

\paragraph{Structure in SAEs.}
SAEs decompose model activations into a high-dimensional set of sparse latent features, and have become a widely used framework for unsupervised concept discovery in large language and vision-language models \citep{bricken2023monosemanticity}. Several post-hoc approaches investigate what structure these feature spaces contain, for example, via decoder geometry \citep{bricken2023monosemanticity, leask2025sparse, li2025geometry}, top-activating inputs \citep{girrbach2025sparse}, textual descriptions \citep{jiang2025interpretable}, or co-activation patterns \citep{bussmann2025learning, ye2024sparse}. \citet{wittenmayer2026insight} explicitly target a hierarchical DAG, using weighted conditional feature probabilities.
Beyond post-hoc analysis, several approaches impose structural constraints during training. \citet{zaigrajew2025interpreting} and \citet{bussmann2025learning} follow the Matryoshka principle using nested sparsity constraints and nested dictionary sizes, respectively, while \citet{li2025unlocking} introduce a non-nested static grouping strategy. \citet{costa2025flat} instead recover hierarchical features by iteratively refining reconstruction residuals, \citet{muchane2025incorporating} enforce hierarchy explicitly by combining a high-level SAE with lower-level expert SAEs, and \citet{luo2026atoms} learn  multiple SAEs and their relations via structural constraints and feature perturbations. 
Not all of these works frame their contribution in hierarchical terms, but each imposes implicit or explicit structure that shapes relationships among learned features. This work introduces a shared framework to assess and compare how well these approaches impose meaningful hierarchies in SAE latent spaces.

\paragraph{SAE Evaluation.}
Standard SAE evaluation focuses on reconstruction quality, sparsity, and feature disentanglement \citep{karvonen2025saebench}, with more recent work also targeting the interpretability of individual features via automated tools \citep{paulo2024automatically} or Monosemanticity scores \citep{pach2025sparse, harle2025measuring}. However, the evaluation of hierarchical structure has received limited attention. Supervised approaches compare against ground-truth hierarchies, typically on small synthetic datasets \citep{bussmann2025learning, costa2025flat, chanin2025feature, chanin2024absorption}. Among unsupervised approaches, \citet{ye2024sparse} evaluate hierarchical strength via parent-child versus child-child co-occurrence ratios, \citet{luo2026atoms} measure hierarchical consistency and conditional co-activation probabilities, and \citet{wittenmayer2026insight} assess extracted feature graphs through a user study; each covering different aspects in isolation. Moreover, existing work relies heavily on qualitative illustrations \citep{bussmann2025learning, muchane2025incorporating, wittenmayer2026insight}, making it difficult to compare methods rigorously. Rather than adding another individual metric, we first establish an explicit, comprehensive set of requirements for what a meaningful hierarchy should satisfy, and then derive a unified, quantitative evaluation protocol from those requirements.

\section{Concept Hierarchies} 
\label{sec:method}

We consider a concept as "the label of a set of things that have something in common" \citep{archer1966psychological}, for example, a \textit{tennis racket} or a \textit{pine cone}, but also more abstract ideas such as \textit{plant}. When reasoning about visual scenes, humans naturally relate concepts and organize them into structures \citep{hafri2021perception}, with hierarchies being a particularly natural choice due to their clear relational semantics. In the following, we formalize this notion and derive the requirements that such hierarchies should satisfy in the context of unsupervised concept learning.

\paragraph{Hierarchical Semantics}
Let $\mathcal{F}$ be a set of concepts used to represent inputs $x \in \mathcal{X}$. We are interested in \textit{generalization/specialization} hierarchies over $\mathcal{F}$, which have a clear semantic relation: for a parent concept $p \in \mathcal{F}$ and a child concept $c \in \mathcal{F}$, $p$ represents a more general property of inputs than $c$ while $c$ captures a more specific property within the scope of $p$. Consequently, the presence of $c$ typically implies the presence of $p$ \citep{brachman1983and}.
This includes typical IS-A relations (tennis is a sport) but also more general relationships (a tennis racket is a specific aspect of tennis).

We model these relations as a \textbf{concept hierarchy graph}, a directed acyclic graph (DAG), defined as $G := (V, E)$ over a subset $\tilde{\mathcal{F}} \subseteq \mathcal{F}$ of all concepts that are part of at least one hierarchical relation, i.e. $V := \tilde{\mathcal{F}}$. For an edge $(p, c) \in E$, $p$ is the parent and $c$ is the child. We assume that $G$ is transitively reduced, meaning an edge $(p, c)$ only exists if there is no directed path of length $\ge 2$ from $p$ to $c$, which avoids redundant ancestry information and keeps the local structure interpretable. The DAG allows concepts with multiple parents, better reflecting the semantics of generalization/specialization hierarchies.
A \textbf{concept extractor} $\mathcal{M}$ maps inputs to a set of feature activations $a(x) = \{a_f(x) \in \mathbb{R}_{\geq 0} | f \in \mathcal{F'}\}$ in its feature space $\mathcal{F'}$. We assume that there is a mapping from $\mathcal{F'}$ to $\mathcal{F}$, and drop this difference to simplify the notation in the following. We also treat $a(x)$ as covering the case where $\mathcal{M}$ operates on latent embeddings $e(x)$ of an encoder model, as in the SAE setting.

We now describe the requirements for concept hierarchies from the perspective of unsupervised concept learning, in particular, requirements for hierarchical concept extractors.\footnote{This assumes a method to obtain a concept hierarchy graph from the concept extractor, either provided inherently \citep{muchane2025incorporating} or constructed post-hoc \citep{wittenmayer2026insight}.}
We write $\mathcal{M}_H$ for a hierarchical concept extractor, $p$ for a parent concept, $c$ for a child concept, and $C(p)$ for the set of children of $p$ in the concept hierarchy graph.

\subsection{Requirements}

\begin{wrapfigure}{r}{0.5\linewidth}
    \vspace{-0.5cm}
    \centering
    \fbox{
    \includegraphics[width=0.95\linewidth]{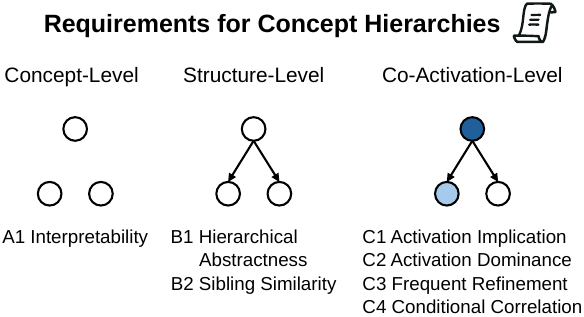}
    }
    \caption{
     \textbf{Requirements for concept hierarchies operate on different graph aspects.} \textit{Concept-level} concerns individual features, \textit{structure-level} addresses the static graph structure, and \textit{co-activation-level} evaluates the hierarchy given sample-based activations.}
    \label{fig:method}
    \vspace{-0.5cm}
\end{wrapfigure}
In this section, we derive requirements that capture the essential properties of concept hierarchies, divided into concept-level, structure-level, and co-activation-level, depending on whether they concern concepts in isolation, the static graph structure, or the dynamic instantiation of that structure across individual samples.

\paragraph{Concept-level.}
Concept-level requirements concern individual concepts in isolation, independent of their position in the hierarchy graph. They represent basic pre-requisites for meaningful hierarchy analysis: studying hierarchical relations is only worthwhile if the underlying concepts satisfy basic quality criteria. 

\textbf{Requirement A1: Interpretability.}
Each node in the graph should correspond to a human-understandable concept from the data $\mathcal{X}$. This is particularly relevant in unsupervised concept discovery, where there is no supervision to guarantee the quality of individual features.

\paragraph{Structure-level.}
These requirements describe desired properties of the hierarchy graph independent of individual samples. Assuming concepts are interpretable, these requirements specify how the position of concepts in $G$ should align with their semantics.

\textbf{Requirement B1: Hierarchical Abstractness.}
The semantics of a \textit{generalization/specialization} hierarchy specify that for an edge $(p, c) \in E$, the child $c$ should be more specific than its parent $p$. For example, \textit{tennis} is more general than \textit{tennis racket} or \textit{tennis serve}. This requirement is either explicitly \citep{brachman1983and} or implicitly \citep{luo2026atoms, muchane2025incorporating, wittenmayer2026insight} assumed in prior work.

\textbf{Requirement B2: Sibling Similarity.}
If a parent $p$ has multiple children, each child should share the concept represented by $p$, forming a coherent set of refinements rather than an arbitrary collection. In our example, all children of \textit{tennis} should be semantically related through their shared parent concept.

\paragraph{Co-activation-level.}
The third level of requirements goes beyond the static hierarchy and considers sample-based activations. In unsupervised concept discovery, it is not sufficient for the hierarchy to encode meaningful semantic relations; the per-sample activation behavior should also reflect the semantics of parent-child relations. 
We treat activations as indicators of the strength of evidence for a concept and consider both discrete (present or absent) and continuous (larger activations equal stronger evidence) settings. Importantly, this level also addresses the problem of absorption \citep{chanin2024absorption}.

\textbf{Requirement C1: Activation Implication.}
Since $c$ is a refinement within the scope of $p$, the presence of $c$ implies the presence of $p$: an active child should always be accompanied by an active parent. This is a standard assumption in hierarchical sparse coding \citep{jenatton2010proximal} and a central property of \textit{generalization/specialization} hierarchies \citep{brachman1983and}. Activation Implication also means that there is no hard form of absorption, i.e., the activation of a child concept does not suppress the activation of its parent concept.
In our example, every input where \textit{tennis racket} is active, \textit{tennis} should also be active.

\textbf{Requirement C2: Activation Dominance.}
When both $p$ and $c$ are active, the activation of the parent should be at least as strong as that of the more specific child. For example, evidence for \textit{tennis serve} should be accompanied by equal or stronger evidence for \textit{tennis}. This need not hold as a hard per-sample constraint: it is possible to be confident about a \textit{racket} without knowing whether it is for \textit{tennis} or \textit{squash}, but it should hold in expectation over the full data distribution. This requirement is strongly related to \textit{activation implication}, refining the activation semantics for an edge $(p, c) \in E$ where both $p$ and $c$ are active.

\textbf{Requirement C3: Frequent Refinement.}
When an abstract parent $p$ is present in an input, it is often not present in its abstract form, but rather in the form of one of its child refinements. This should also be reflected in the co-activations: an active parent $p$ should frequently be accompanied by at least one active child $c \in C(p)$. While refinement frequency is desired to be high, it is not a universal constraint: a parent may appear without an active child if the relevant specialization is not contained in $C(p)$ or if the parent is present in a generic form that needs no further refinement. In our example, an image containing a \textit{tennis racket} should activate not only \textit{tennis} but also \textit{tennis racket}.

\textbf{Requirement C4: Conditional Correlation.}
While \textit{activation dominance} concerns only the binary comparison of parent-child pairs, this requirement takes a broader view: the activation magnitude of $p$ should \textit{correlate} positively with the activation magnitudes of its active children in $C(p)$, given that at least one child is active. This directly addresses a "soft" form of absorption, in which stronger child activations gradually suppress the parent rather than triggering a hard activation drop. In our example, across all samples where at least one child of \textit{tennis} is active, the activation strength of \textit{tennis} should correlate with that of its active children, regardless of which children are active in any given sample.

\subsection{Metrics}
We now provide concrete metrics to measure each requirement. We assume a concept extractor with a corresponding concept hierarchy graph, but do not assume access to a ground-truth hierarchy. Therefore, all metrics evaluate requirements in an unsupervised manner. Depending on the requirement, metrics operate at the feature level, the parent-child level, or the parent-child group level (cf.\ \autoref{tab:desiderata_metric_overview}).

\textbf{Notation.} Given a dataset $D$, we write $\mathbb{P}_d[\cdot]$ for the empirical probability over $D$. Given a (hierarchical) concept extractor $\mathcal{M}_H$, each feature $f \in V$ induces an activation $a_f(x) \in \mathbb{R}_{\geq 0}$, with its binarized form $A_f(x) \in \{0, 1\}$ which indicate whether the feature $f$ is considered \textit{active} on input $x$. We write $\mathrm{TopKInputs}_f := \{x^{(1)}, \dots, x^{(k)}\}$ for the $k$ inputs in $D$ with the largest activations under $a_f$, and $\csim(u,v):= \frac{\langle u,v\rangle}{\|u\|\,\|v\|}$ for cosine similarity.

\textbf{Monosemanticity Score $\texttt{s}_m$.}
Concept-level interpretability has been extensively studied in the SAE literature, with a range of metrics and evaluation protocols available \citep{paulo2024automatically}; we therefore do not introduce a dedicated interpretability metric here, but refer to related work for such comparisons \citep{shu2025survey, karvonen2025saebench}. Instead, we use the Monosemanticity Score, following \citet{pach2025sparse}, which evaluates semantic coherence in an unsupervised manner and provides an important baseline perspective for the hierarchical metrics below. Concretely, it computes the average pairwise similarity between input embeddings, weighted by feature activations.
Let $\tilde{a}_f(x)$ be the min-max normalized activations of $f$ on $x$, and $e(x)$ the latent embeddings of $x$ from an input encoder, then $\texttt{s}_m \in [-1, 1]$ is  
\begin{equation}
    \texttt{s}_{m}(f) := \frac{\sum_{x^{(i)}, x^{(j)} \in D}\,\, \tilde{a}_f(x^{(i)})\,\, \tilde{a}_f(x^{(j)})\, \csim\bigl(e(x^{(i)}), e(x^{(j)})\bigr)}{\sum_{x^{(i)}, x^{(j)} \in D}\,\, \tilde{a}_f(x^{(i)})\,\, \tilde{a}_f(x^{(j)})}\,.
\end{equation}

\textbf{Hierarchical Abstractness Score $\texttt{s}_{ha}$.} 
This metric is computed per parent-child pair and uses the Monosemanticity Score as a proxy for specificity, treating specificity as the inverse of abstractness: a high $\texttt{s}_m$ indicates a more specific feature, while a low suggests a more abstract concept. 
Let $(p,c)\in E$ be a parent-child edge in the graph and $\mathbf{1}[\cdot]$ the indicator function. We define the binary 
\begin{equation}
\texttt{s}_{ha}(p,c):=\mathbf{1}\!\left[\texttt{s}_m(p) < \texttt{s}_m(c)\right].
\end{equation}

\textbf{Sibling Similarity Score $\texttt{s}_{ss}$.}
This metric measures the similarity among the children $c \in C(p)$ of a parent $p$, using cosine similarity between activation-weighted averages of input encoder latents, inspired by \citet{dreyer2025mechanistic}. For a parent $p\in V$, the $\texttt{s}_{ss} \in [-1, 1]$ is defined as:
\begin{equation}
\texttt{s}_{ss}(p)
 := \frac{2}{|\mathcal{C}(p)|\,(|\mathcal{C}(p)|-1)}\sum_{c_i,c_j\in \mathcal{C}(p); i<j}
\csim\!\left(\bar{e}_{f_i},\bar{e}_{f_j}\right)\,,
\end{equation}
with $\bar{e}_f := \frac{\sum_{x\in \mathrm{TopKInputs}_f} a_f(x)\,e(x)}{\sum_{x\in \mathrm{TopKInputs}_f} a_f(x)}$. This metric is only defined for $|\mathcal{C}(p)|\ge 2$.

\textbf{Activation Implication Score $\texttt{s}_{ai}$.} We measure how often a parent is active given an active child via the conditional activation probability on binarized activations, similar to  \citep{luo2026atoms}. For $(p,c)\in E$:
\begin{equation}
\texttt{s}_{ai}(p,c) := \mathbb{P}_D\bigl[A_p(x) = 1 \,\big|\, A_c(x) = 1\bigr].
\end{equation}

\textbf{Activation Dominance Score $\texttt{s}_{ad}$.}
This metric measures the fraction of inputs on which the parent activation exceeds the child activation, conditioned on both features being active. Conditioning on the child being active prevents the metric from being dominated by how often a child is active relative to its parent. Conditioning on the parent being active excludes cases that are already captured by \textit{activation implication}, i.e., where the parent is inactive while the child is active. For $(p,c)\in E$:
\begin{equation}
\texttt{s}_{ad}(p,c)
:= \mathbb{P}_D\!\left[a_p(x)\geq a_c(x)\,\middle|\, (A_p(x) = 1) \wedge (A_c(x) = 1)\right].
\end{equation}

\textbf{Refinement Frequency Score $\texttt{s}_{rf}$.} This metric measures how often at least one of a parent's children is active, given that the parent is itself active. As we do not assume specific dependencies between siblings, we consider only the presence of at least one active child rather than the total number of active children. While it is desirable that a parent is often accompanied by its children, this does not always have to be the case, as discussed for the respective requirement. For $p\in V$:
\begin{equation}
\texttt{s}_{rf}(p)
:= \mathbb{P}_D\!\left[
\sum\nolimits_{c\in \mathcal{C}(p)} A_c(x)\ge 1
\,\middle|\,
A_p(x) = 1
\right].
\end{equation}

\textbf{Conditional Correlation Score $\texttt{s}_{cc}$.} 
This metric measures the Spearman correlation $\rho_s$ between activations of a parent feature and the maximum activation of any of its children. It is conditioned on at least one child being active to measure correlations among active features, rather than how often a child is active relative to the parent. We use Spearman rather than Pearson correlation as we are interested in a monotonic rather than strictly linear relationship. For $p\in V$, the $\texttt{s}_{cc} \in [-1, 1]$ is:
\begin{equation}
\texttt{s}_{cc}(p)
:= \rho_s\Bigl(\{a_p(x)\}_{x\in D_p},\, \{\max_{c\in \mathcal{C}(p)} a_c(x)\}_{x\in D_p}\Bigr)\,,
\end{equation}
with $D_p := \Big\{x \in D : \sum_{c\in \mathcal{C}(p)} A_c(x) \ge 1\Big\}$.

\paragraph{Metric Aggregation.} Graph-level scores for each metric are obtained by averaging local scores. Thereby, parents contribute proportionally to the number of their children for parent-child metrics, whereas each parent contributes equally for parent-child-group metrics.

\section{Experimental Evaluation} \label{sec:experiments}

In this section, we evaluate the current state of SAEs with respect to their hierarchical feature representations, comparing standard TopK SAEs against several hierarchical approaches. We aim to answer the following research questions: (1) What is the current state of hierarchical visual concept learning in SAEs? (2) How does the choice of vision encoder affect hierarchy quality? (3) How does hierarchy quality vary with depth? Let us first describe the experimental setup.

\textbf{SAE Approaches.} We compare six SAE types: a standard TopK SAE \citep{bricken2023monosemanticity}, a Matryoshka SAE with nested sparsity constraints (ActMSAE) \citep{zaigrajew2025interpreting}, a Matryoshka SAE with nested dictionary sizes (ArchMSAE) \citep{bussmann2025learning}, EWG-SAE \citep{li2025unlocking}, Matching-Pursuit SAE (MP-SAE) \citep{costa2025flat}, and H-SAE \citep{muchane2025incorporating}. Hyperparameter details for each method are provided in \autoref{app:sae_details}.

\textbf{Data and Encoders.} All SAE approaches are trained on global image embeddings from CLIP ViT-L/14 \citep{radford2021learning}, DINOv2-base \citep{oquab2023dinov2} as well as patch-level residual stream encodings of Gemma-4 E2B \citep{team2026gemma} over CC3M \citep{sharma2018conceptual}. The dataset is divided into three disjoint splits: graph creation (500,000 samples), evaluation (500,000 samples), and the rest for SAE training.

\textbf{Graph Creation.} Evaluating our metrics requires a concrete concept hierarchy graph. Since most hierarchical SAE approaches do not produce an explicit graph but only impose implicit structural constraints during training, we follow \citet{wittenmayer2026insight} and construct a DAG from pairwise conditional activation probabilities between features. We apply several post-processing steps to the resulting graph: removing noise by discarding overly small components and overly dense nodes, followed by a transitive reduction to eliminate redundant edges. Further details are given in \autoref{app:graph_creation_details}.

\begin{table}[t]
\centering
\caption{\textbf{SAEs learn mostly shallow concept hierarchies.} The graph-level statistics for SAEs trained on CLIP highlight different graph sizes and densities, while all methods except ActMSAE are mostly limited to two-level hierarchies. Values are mean $\pm$ std across three seeds.  \\
\footnotesize$^\dagger$Group Size Deviation is the standard deviation of the per-seed mean group size across seeds.}
\label{tab:clip_graph_metrics}
\resizebox{\textwidth}{!}{
\begin{tabular}{@{}lrrrrrrr@{}}
\toprule
\textbf{Model}
  & \textbf{Nodes}
  & \textbf{Edges}
  & \textbf{Parent-Child}
  & \textbf{Max}
  & \textbf{Avg. Group}
  & \textbf{Group Size}
  & \textbf{Nodes at} \\
& & & \textbf{Groups} & \textbf{Depth} & \textbf{Size} & \textbf{Deviation}$^\dagger$ & \textbf{Depth $\geq 3$} \\
\midrule
TopK SAE  & $2120 \pm 42$  & $3011 \pm \phantom{0}77$  & $225 \pm 16$  & $3.67 \pm 0.58$ & $13.40 \pm 0.92$ & $19.03 \pm 1.64$ & $80 \pm 15$   \\
EWGSAE   & $974 \pm 57$   & $2641 \pm 192$ & $168 \pm 10$  & $2.67 \pm 0.58$ & $15.77 \pm 1.09$ & $22.14 \pm 2.04$ & $3 \pm \phantom{0}3$     \\
ArchMSAE & $1012 \pm 99$  & $1683 \pm 190$ & $99 \pm \phantom{0}2$    & $3.00 \pm 0.00$ & $16.98 \pm 1.61$ & $19.16 \pm 6.37$ & $2 \pm \phantom{0}1$     \\
ActMSAE  & $3341 \pm 19$  & $6170 \pm \phantom{0}81$  & $1372 \pm 14$ & $5.67 \pm 0.58$ & $4.50 \pm 0.09$  & $8.99 \pm 1.07$  & $1147 \pm 29$ \\
H-SAE  & $6137 \pm \phantom{0}0$   & $5776 \pm \phantom{0}\phantom{0}0$   & $361 \pm \phantom{0}0$   & $2.00 \pm 0.00$ & $16.00 \pm 0.00$ & $0.00 \pm 0.00$  & $0 \pm \phantom{0}0$     \\
\bottomrule
\end{tabular}
}
\end{table}

\paragraph{Evaluating Learned Concept Hierarchies.}
We first compare the general graph structures across methods (cf.\ \autoref{tab:clip_graph_metrics} for CLIP, \autoref{app:further_graph_level} for further details). Post-hoc constructed graphs cover up to half of all SAE features, with MP-SAE being the only exception: its training strategy reduces co-activation of similar features (\autoref{app:graph_creation_details}), resulting in an empty graph. The explicit hierarchy of H-SAE, by contrast, always contains all 6,137 features. Notably, ActMSAE is the only method producing a hierarchy with more than two levels, suggesting better co-activation across multiple levels of abstraction. Graph statistics are stable across seeds, indicating that the learned structures are not sensitive to the SAE training process.

\begin{figure}
    \centering
    \includegraphics[width=0.88\linewidth]{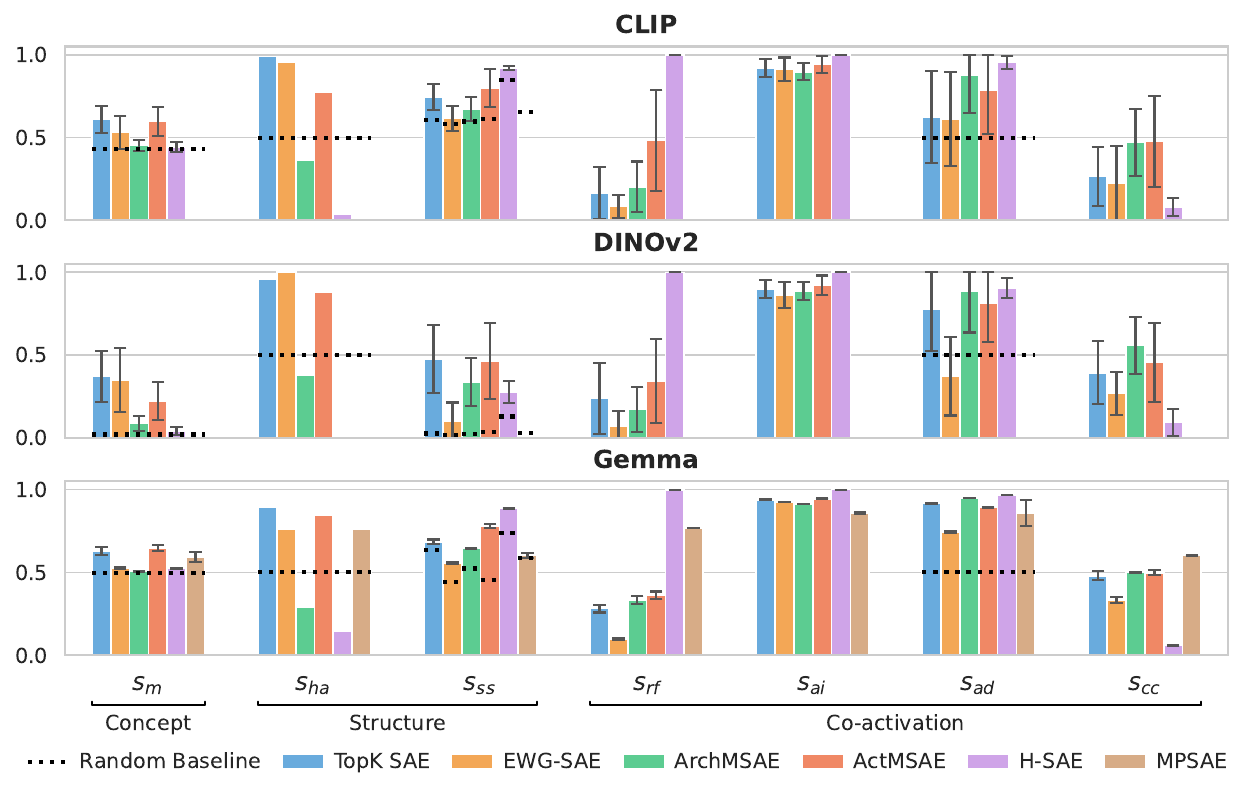}
    \caption{\textbf{Graph metrics for all SAEs trained on CLIP, DINOv2, and Gemma-4 encodings.} Average and standard deviation over nodes, edges, or parent-child groups. Random baseline uses a random selection of respective samples/nodes/edges where applicable. Full data distributions (cf. \autoref{app:more_qualitative}) show that values $< 0$ can occur for $\texttt{s}_{cc}$. As $\texttt{s}_{ha}$ is binary, we omit its standard deviation.
    }
    \label{fig:clip_results}
\end{figure}

Turning to the graph metrics (cf.\ \autoref{fig:clip_results}), several clear trends emerge. The baseline TopK SAE already produces features that can form a meaningful concept hierarchy: the extracted graph covers roughly one-third of all features and achieves strong abstractness and sibling similarity scores. However, its co-activation metrics reveal that absorption remains a key challenge. Even when hard absorption is controlled for via the graph construction threshold, TopK SAE shows low $\texttt{s}_{ad}$ and $\texttt{s}_{cc}$, meaning parent and child activations are largely inconsistent in strength. The low $\texttt{s}_{rf}$ further confirms that, while features are semantically grouped, children rarely co-activate with their parents.

EWG-SAE shows no meaningful improvement over the baseline and often scores slightly worse. The Matryoshka approaches perform better on co-activation metrics overall, with ActMSAE generally exceeding ArchMSAE. ArchMSAE predominantly learns abstract features with no clear hierarchical abstraction (as indicated by around random $\texttt{s}_m$ and $\texttt{s}_{ha}$), suggesting that nested dictionary sizes reduce absorption but constrain co-activations to rather abstract features. ActMSAE, with its nested sparsity constraints, instead produces more frequent and consistent co-activations of specific features with their general parents, as reflected in its high $\texttt{s}_{rf}$ and multi-level graph. 

The explicit hierarchy of H-SAE yields consistent co-activation behavior and high refinement frequency by design, but soft absorption persists: parent and child activation strengths remain poorly correlated. Moreover, its features achieve near-random $\texttt{s}_{m}$ and near-zero $\texttt{s}_{ha}$, indicating that the imposed hierarchy lacks semantic meaning, as children are more abstract than their parents or contain no clear interpretation (cf.\ \autoref{app:more_qualitative}). Thus, explicit structural constraints can enforce co-activation consistency, but do not reduce the necessity to maintain clear semantic structure within the hierarchy.

\paragraph{Comparing Hierarchies Across Vision Encoders.}
To investigate the impact of the encoder on hierarchical feature learning, we compare SAEs trained on CLIP embeddings with those trained on DINOv2 and Gemma-4 E2B. In contrast to CLIP's contrastive vision-language pretraining, DINO is trained self-supervised on vision data alone, and for Gemma-4, we train SAEs on the residual stream of patch-level image encodings. Differences between encoders include the substantially lower average similarity between image embeddings for DINOv2 (${\approx}0.01$), which results in smaller graphs on average, and that MP-SAE produces a non-empty graph when trained on Gemma-4 E2B, albeit consisting only of one parent-child group (cf. \autoref{tab:gemma_graph_metrics}). The otherwise stable trends across encoders are a strong indicator that, as long as the embeddings contain enough information to learn suitable hierarchies, the challenges of learning hierarchical concept spaces are driven by SAE training strategies rather than the underlying embedding space.

\begin{figure}
    \centering
    \includegraphics[width=\linewidth]{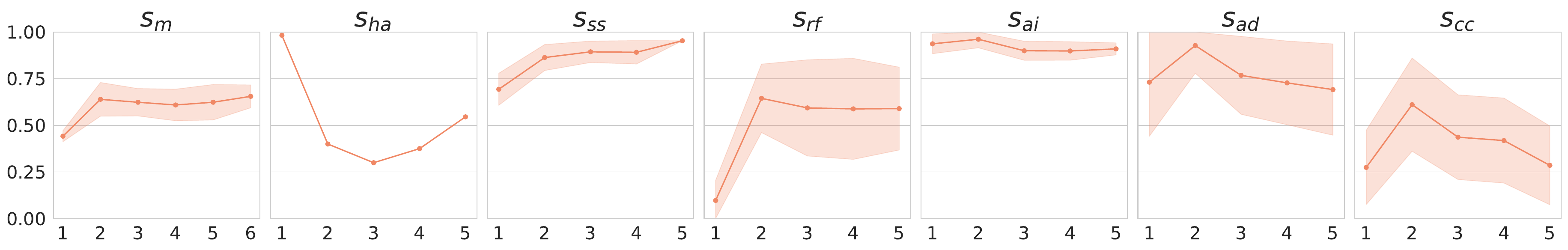}
    \caption{\textbf{Hierarchy quality of ActMSAE across depths.} While the hierarchy abstraction diminishes at higher depth, semantic coherence among children and refinement frequency increase. Values are aggregated at the depth level of the node ($\texttt{s}_m$) or of the parent for edge and group metrics.}
    \label{fig:depth_analysis}
\end{figure}

\paragraph{Investigating More Than Two-Level Hierarchies.}
Since all methods except ActMSAE produce shallow two-level hierarchies, we use ActMSAE on CLIP to investigate how hierarchy quality changes with depth (cf.\  \autoref{fig:depth_analysis}). The $\texttt{s}_{m}$ and $\texttt{s}_{ha}$ values reveal a clear difference in feature specificity between levels 1 and 2, while deeper levels remain similar. Refinement frequency increases with depth: top-level abstract features rarely co-activate with their children, while more specific features co-occur much more frequently. The remaining co-activation metrics show no clear trend, suggesting that absorption behavior is largely independent of hierarchy depth. Overall, ActMSAE produces a deeper graph with a meaningful abstraction gap between the first two levels, and lower levels consist of frequently co-occurring features with increasingly subtle hierarchical relations.

\begin{figure}
    \centering
    \includegraphics[width=\linewidth]{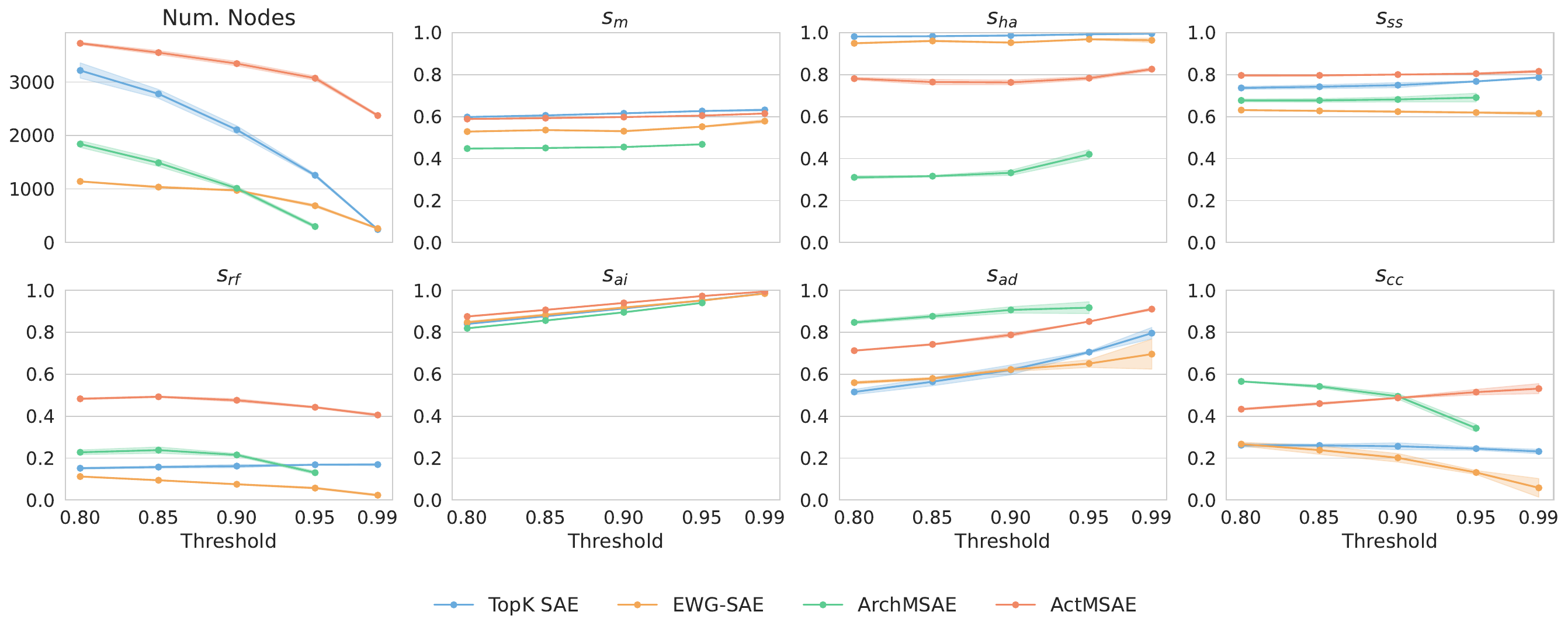}
    \caption{\textbf{Ablation of the graph-creation threshold on CLIP.} Mean $\pm$ standard deviation over three seeds. While the threshold on conditional co-activation affects graph size and the co-activation-level metrics, the relative comparisons between SAE types remain largely consistent.}
    \label{fig:graph_treshold_ablation}
\end{figure}

\paragraph{Ablations on Graph Creation.}
Since the results for most SAE types also depend on the method used to build the hierarchy graph, we ablate the graph-creation threshold, the graph post-processing, and the similarity measure used for graph creation. 
Varying the threshold on conditional activations (\autoref{fig:graph_treshold_ablation}) mainly affects $\texttt{s}_{ai}$, $\texttt{s}_{cc}$, and graph size, which drops noticeably at very high thresholds, but the relative comparison between SAE methods remains stable. 
The post-processing steps improve human readability (transitive reduction) or hierarchy quality (graph pruning) without altering the main findings (cf. \autoref{app:graph_creation_ablation}).
Finally, replacing conditional co-activations (CCA) with masked cosine similarity (MCS) preserves the main trends, though MCS-based graphs show lower $\texttt{s}_{ai}$, $\texttt{s}_{ad}$, and $\texttt{s}_{cc}$. MCS does allow constructing a graph for MPSAE for CLIP, but this graph has near-random $\texttt{s}_m$ and near-zero $\texttt{s}_{ha}$, indicating that its hierarchy lacks semantic meaning (cf. \autoref{tab:graph_metric_ablation} in the appendix).

\begin{figure}
    \centering
    \begin{subfigure}[t]{0.6\linewidth}
        \centering
        \includegraphics[height=3.0cm]{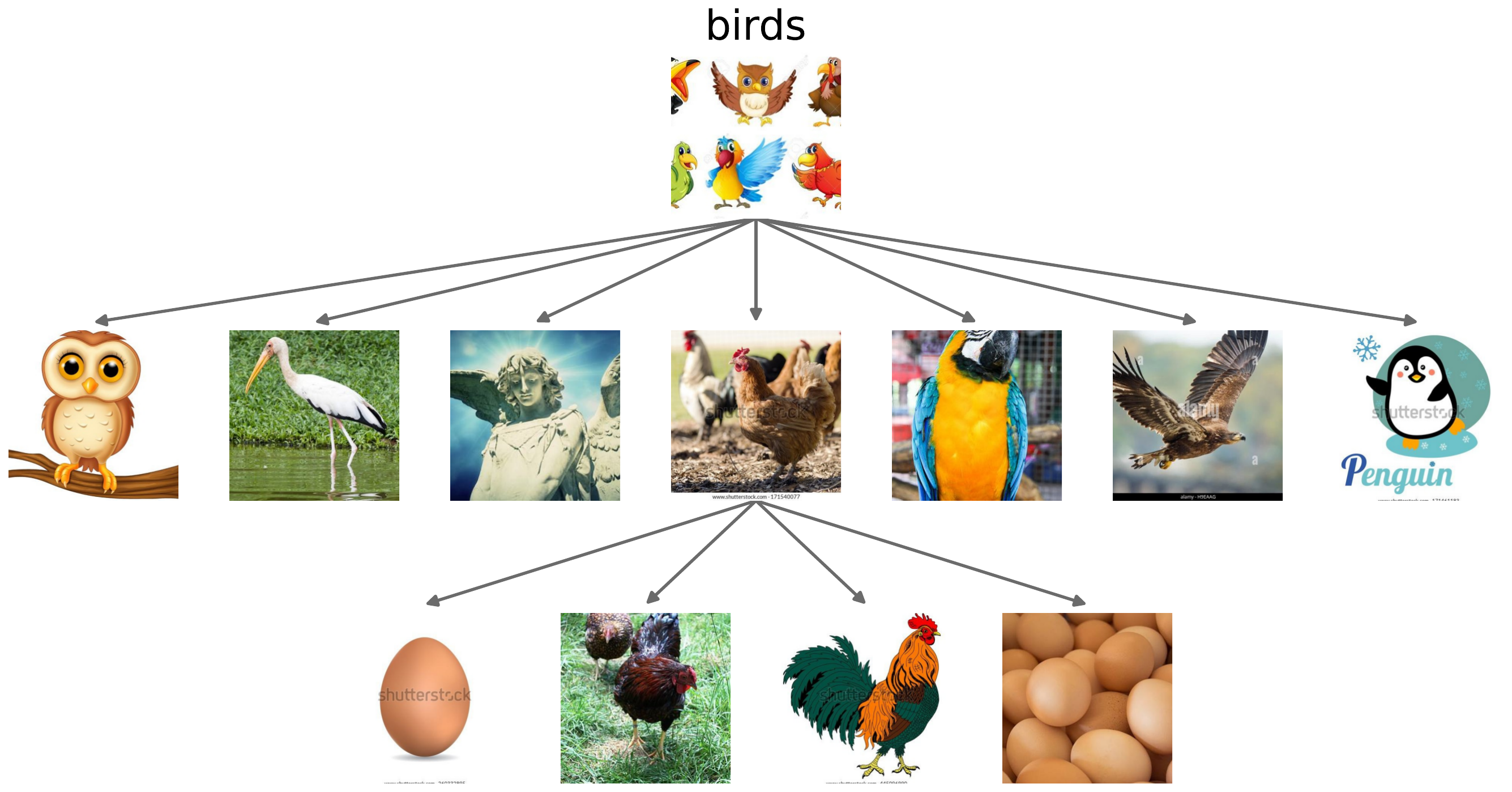}
    \end{subfigure}
    \hfill
    \begin{subfigure}[t]{0.3\linewidth}
        \centering
        \includegraphics[height=3.0cm]{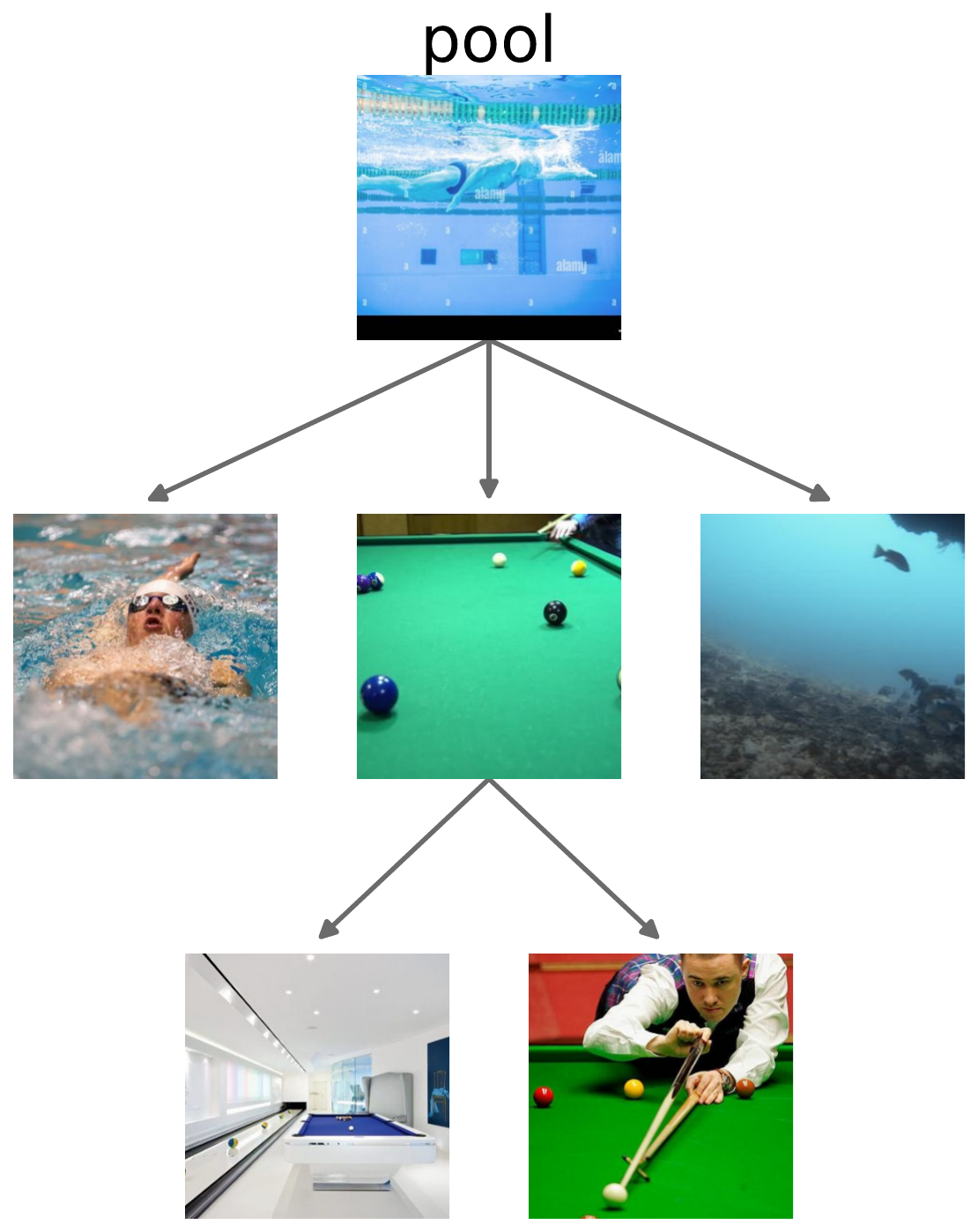}
    \end{subfigure}
    \caption{\textbf{Learned hierarchies capture meaningful semantic abstractions.} Left: various bird types are grouped under a common \textit{birds} concept, with chickens further refined into roosters, hens and eggs. Right: CLIP's vision-language pretraining shapes the learned hierarchies, causing swimming pools and billiards tables to be grouped under the shared concept \textit{pool}. All examples taken from ActMSAE trained on CLIP.}
    \label{fig:qualitative}
\end{figure}

\paragraph{Qualitative Hierarchy Analysis.}
We complement the quantitative evaluation with qualitative examples of learned hierarchies. The first (cf.\ \autoref{fig:qualitative}, left) shows a dense hierarchy grouping various bird types and their specializations. Interestingly, angels are also grouped as birds, likely due to their wings and feathers.
Chicken are further refined via roosters, hens and eggs, which highlights a sensible more than two-level categorization. On the other hand, this also illustrates that child activation may sometimes exceed parent activation in a sensible way. For samples with an egg, it can very well be that the model is more certain about the egg than about whether it is an egg of a chicken, highlighting why the activation dominance and conditional correlation metrics should be viewed in expectation over the dataset instead of hard per-sample constraint.

The second example illustrates a more interesting case: the top-level concept \textit{pool} is refined into swimming pools, fishing ponds, and billiards, with billiards further split into two subcategories. This grouping reflects CLIP's language supervision, which places visually distinct concepts close together due to the shared word \textit{pool}. This highlights how the choice of encoder can directly shape the hierarchical connections that emerge, a factor quantitative metrics alone do not fully capture. On the other hand, there are two typical failure modes of learned concept hierarchies: grouping only spuriously co-activating features and the presence of noisy features that lack a consistent underlying concept (cf.\ \autoref{app:more_qualitative} for concrete examples of these failure modes).

\section{Discussion} \label{sec:discussion}
Our evaluations show that SAE feature spaces generally contain features that can be grouped into meaningful hierarchies, but obtaining consistent activation behavior remains difficult. A central source of this difficulty is feature absorption, which is typically discussed in its "hard" form, i.e., whether a feature is active or not \citep{chanin2024absorption}. Our evaluation reveals a complementary "soft" form: when both parent and child are active, stronger child activations tend to suppress parent activations, as captured by \textit{activation dominance} and \textit{conditional correlation}. Matryoshka approaches reduce hard absorption, but remain susceptible to this soft form.
Addressing both forms of absorption might require rethinking the sparsity constraints central to current SAE training. A less sparse SAE with consistent hierarchical activations could be more useful than a sparser one with no meaningful structure.
A second open question is whether hierarchical structure should be explicitly enforced during training or implicitly induced and recovered post-hoc. Explicit constraints can guarantee that certain requirements are met, while post-hoc graph construction adapts more flexibly to the features the SAE has learned. A promising direction likely combines both: fixing a high-level hierarchical structure with clear activation semantics, while discovering the lower parts of the hierarchy flexibly.

In general, our metrics matter differently depending on the use case. For understanding the learned concept hierarchy itself (exploring the concept space), co-activation metrics are less important while structure- and concept-level metrics are the focus.
For steering, relevance depends on the scenario: if the target concept and its parent are already known to be semantically sensible, co-activation metrics matter most. If this is not yet established, structure- and concept-level metrics are needed first to verify the hierarchy is meaningful.
More generally, these metrics build on each other: trusting the structure-level metrics assumes that the concept-level metrics are reasonable, and trusting the co-activation metrics assumes that the structure-level metrics are reasonable in turn.

\textbf{Limitations.} By design, all metrics in this work operate unsupervised, approximating, for example, semantic similarity through image embeddings rather than human judgment. 
While this enables scalable evaluation, and we verify the structure-level metrics against the known, ground-truth ImageNet-1k hierarchy \citep{deng2009imagenet} in \autoref{tab:imagenet_hierarchy_eval}, this set of metrics does not fully capture whether a hierarchy is meaningful to humans. Thus, quantitative metrics and human inspection are best understood as complementary tools rather than substitutes for one another.
A second limitation concerns the dependency on the graph construction method of \citet{wittenmayer2026insight} for all approaches except H-SAE. Since this method considers only conditional activation probabilities, it captures just one of the hierarchy requirements and may not yield the best possible graph for every SAE.
More broadly, how to best extract a hierarchy graph from SAE latent representations remains an open question that warrants further investigation.

\section{Conclusion} \label{sec:conclusion}

In this work, we derived a general set of requirements for concept hierarchies for unsupervised concept learning, covering concept-level prerequisites alongside structure- and co-activation-level requirements. From these, we developed a concrete set of metrics for evaluating SAE-based concept hierarchies. Our evaluation of current hierarchical SAE approaches shows that they can discover features that can be grouped into semantically meaningful, although mostly shallow, hierarchies. However, activation behavior remains inconsistent across hierarchy levels, and absorption emerges as a central challenge, both in its previously known hard form and in a newly identified soft form, where stronger child activations suppress parent activations even when both are active.
Looking ahead, the proposed requirements and metrics can serve as direct training constraints for future hierarchical concept extractors. Open questions remain how to best extract hierarchy graphs from SAE latent spaces, and how much hierarchical structure should be explicitly enforced during training. Applying this evaluation protocol to other modalities or concept extractors beyond SAEs is a natural next step towards a better understanding of unsupervised hierarchical concept learning.

\section*{Acknowledgements}
This work was supported by the ”ML2MT” project from the Volkswagen Stiftung, by the German Research Foundation (DFG) under Germany’s Excellence Strategy (EXC 3066/1 “The Adaptive Mind”, Project No. 533717223 and by the EU-funded “TANGO” project (EU Horizon 2023, GA No 57100431).
It has benefited from the HMWK project Hessian.AI, and from the Cluster of Excellence "Reasonable AI" funded by the DFG under Germany’s Excellence Strategy EXC-3057.

\bibliography{main}
\bibliographystyle{plainnat}

\appendix

\section{Supplementary material}

\subsection{Broader Impact} \label{app:impact}
This work addresses the current state of hierarchical representations in unsupervised concept learning. By establishing general requirements for concept hierarchies and concrete metrics for evaluating hierarchical SAE approaches, we aim to support the development of more interpretable and structured unsupervised concept representations. We note, however, that the metrics introduced here are not a substitute for careful human inspection of discovered concepts — they are best understood as a complementary tool that can guide and inform, but not replace, human judgment when working with unsupervised concept discovery.

\subsection{Compute Ressources}\label{app:compute}
The experiments in this work have been run on an Nvidia RTX A6000 GPU with 48GB VRAM. Training the different SAE approaches took approximately 65 GPU-hours. The evaluation of the SAEs based on the hierarchy metrics took around 225 GPU-hours, with the largest amount of compute used for the Monosemanticity Score $\texttt{s}_m$, which requires computing pairwise similarities between image embeddings across the dataset.

\section{Details on Training and Experimental Setup}\label{app:experimental_setup}

\subsection{Gemma-4 Implementation details}
We train the SAEs on the residual stream of the frozen Gemma-4 E2B multimodal decoder \citep{team2026gemma}, read at layer~26 of~35 at image patch-token positions ($d=1536$). Each image is pushed through the vision encoder and connector with no text prompt, and we keep the hidden states at the 280 image-token positions. Gemma-4's image processor treats its soft-token budget as an upper bound and produces a per-image grid shape, so we first map every image onto a fixed $960 \times 672$ canvas (resize to cover, then center crop), which yields a fixed $14 \times 20 = 280$ soft-token grid. Therefore, the SAEs trained on Gemma-4 produce feature activations at the patch level rather than a global-image embedding like CLIP and DINOv2. Because each image is represented by 280 patches, we use 50000 images each for SAE training and graph construction, and 70000 for hierarchy evaluation. All SAEs have $6144$ latents, use $k=64$ where the architecture is top-$k$, and are trained for 10 epochs with three seeds.

\subsection{Standard SAE Evaluation Metrics}
We report standard evaluation metrics for SAEs in \autoref{tab:base_metrics_clip}, \autoref{tab:base_sae_metrics_dino} and \autoref{tab:base_sae_metrics_gemma}. Following \citet{zaigrajew2025interpreting}, we report fraction of variance unexplained (FVU)\footnote{FVU is computed as $\frac{1}{N}\sum_{i=1}^{N} \frac{\text{Var}(x^{(i)} - \hat{x}^{(i)})}{\text{Var}(x^{(i)})}$, where $N$ is the number of datapoints.}, cosine similarity, and $L_0$. Both FVU and cosine similarity are used to assess reconstruction quality.
Note that $L_0$ is effectively constant as we use TopK with $k=64$ for all SAEs to calculate these metrics (as all activations within TopK are greater than 0, which is usually the case). The scores are calculated on the validation split of the CC3M~\citep{sharma2018conceptual} dataset.

\subsection{SAE Implementation Details}
\label{app:sae_details}

\paragraph{Embedding preprocessing}
Foundation-model embeddings were mean-centered and normalized following the preprocessing used in \citet{zaigrajew2025interpreting}.

\paragraph{SAE Hyperparameters}
For TopK SAE, ActMSAE, ArchMSAE, EWG-SAE, the encoder subtracts the decoder bias $b_d$ from the input before encoding, uses an encoder bias $b_e$, and adds $b_d$ back after reconstruction (following \citep{zaigrajew2025interpreting}):
\[
f = \sigma\!\left(W_{\mathrm{e}}(x - b_{\mathrm{d}}) + b_{\mathrm{e}}\right),
\qquad
\hat{x} = W_{\mathrm{d}} f + b_{\mathrm{d}}.
\]
Only MP-SAE does not use $b_{\mathrm{e}}$ due to its iterative reconstruction procedure. H-SAE does not use any bias terms, i.e., $b_d = 0$ and $b_e = 0$ following \citep{muchane2025incorporating}.

All SAE variants operate on foundation-model embeddings of dimension $768$ ($1536$ for Gemma-4). To make the models comparable, we use a latent width of $6144$ for all models except H-SAE, which has $6137$ total features due to its expert-based architecture. Unless stated otherwise, encoder and decoder weights are untied, the activation function is ReLU, and models with an explicit fixed sparsity constraint use $L_0=64$. MP-SAE is the only model with tied weights and does not use an activation function. H-SAE uses LeakyReLU instead of ReLU. Table~\ref{tab:sae-specific-hparams} lists the model-specific architectural hyperparameters.

\begin{table}[t]
    \centering
    \caption{Base SAE Metrics for CLIP}
        \begin{tabular}{lllll}
        \toprule
         & FVU & Cosine Similarity & $L_0$ \\
        \midrule
        TopK SAE & 0.1104 ± 0.0009 & 0.9441 ± 0.0003 & 64 ± 0.0000 \\
        EWG-SAE & 0.1362 ± 0.0004 & 0.9291 ± 0.0002 & 64 ± 0.0000 \\
        ArchMSAE & 0.1071 ± 0.0003 & 0.9448 ± 0.0002 & 64 ± 0.0000 \\
        ActMSAE & 0.1120 ± 0.0001 & 0.9421 ± 0.0001 & 64 ± 0.0000 \\
        MP-SAE & 0.0609 ± 0.0000 & 0.9690 ± 0.0000 & 64 ± 0.0000 \\
        H-SAE & 0.1145 ± 0.0001 & 0.9407 ± 0.0001 & 64 ± 0.0000 \\
        \bottomrule
        \end{tabular}
    \label{tab:base_metrics_clip}
\end{table}

\begin{table}[t]
    \centering
    \caption{Base SAE Metrics for DINO}
        \begin{tabular}{lllll}
        \toprule
         & FVU & Cosine Similarity & $L_0$ \\
        \midrule
        TopK SAE & 0.2029 ± 0.0007 & 0.8950 ± 0.0002 & 64 ± 0.0000 \\
        EWG-SAE & 0.3037 ± 0.0002 & 0.8342 ± 0.0002 & 64 ± 0.0000 \\
        ArchMSAE & 0.2217 ± 0.0002 & 0.8823 ± 0.0001 & 64 ± 0.0000 \\
        ActMSAE & 0.2320 ± 0.0003 & 0.8771 ± 0.0002 & 64 ± 0.0000 \\
        MP-SAE & 0.1418 ± 0.0002 & 0.9263 ± 0.0001 & 64 ± 0.0000 \\
        H-SAE & 0.2647 ± 0.0003 & 0.8560 ± 0.0002 & 64 ± 0.0000 \\
        \bottomrule
        \end{tabular}
    \label{tab:base_sae_metrics_dino}
\end{table}

\begin{table}[t]
    \centering
    \caption{Base SAE Metrics for Gemma-4}
        \begin{tabular}{lllll}
        \toprule
         & FVU & Cosine Similarity & $L_0$ \\
        \midrule
        TopK SAE & 0.1920 ± 0.0002 & 0.8985 ± 0.0001 & 64 ± 0.0000 \\
        EWG-SAE  & 0.1850 ± 0.0001 & 0.9048 ± 0.0001 & 64 ± 0.0000 \\
        ArchMSAE & 0.2153 ± 0.0004 & 0.8853 ± 0.0003 & 64 ± 0.0000 \\
        ActMSAE  & 0.2053 ± 0.0001 & 0.8910 ± 0.0000 & 64 ± 0.0000 \\
        MP-SAE   & 0.1450 ± 0.0025 & 0.9244 ± 0.0013 & 64 ± 0.0000 \\
        H-SAE    & 0.2168 ± 0.0003 & 0.9452 ± 0.0001 & 64 ± 0.0000 \\
          \bottomrule
        \end{tabular}
    \label{tab:base_sae_metrics_gemma}
\end{table}

\begin{table}[h]
\centering
\caption{Model-specific architectural hyperparameters.}
\label{tab:sae-specific-hparams}
\renewcommand{\arraystretch}{1.25}
\begin{tabular}{p{0.18\textwidth}p{0.74\textwidth}}
\hline
Model & Hyperparameters \\
\hline

TopK SAE &
\parbox[t]{0.74\textwidth}{
TopK with $K=64$.
}
\\

EWG-SAE &
\parbox[t]{0.74\textwidth}{
Non-nested group sizes: $[64,128,256,512,1024,2048,2112]$. \\
Global sparsity weight: $1.0$.\\
Group sparsity weights proportional to \\
$[0.5^7,0.5^6,0.5^5,0.5^4,0.5^3,0.5^2,0.5^1]$.\\
Group sparsity weights renormalized to sum to $1$.
}
\\

ArchMSAE &
\parbox[t]{0.74\textwidth}{
TopK with $K=64$.\\
Nested widths: $M=\{96,288,672,1440,2976,6144\}$.
}
\\

ActMSAE &
\parbox[t]{0.74\textwidth}{
TopK levels: $k_1,\ldots,k_8=\{64,128,256,512,1024,2048,4096,6144\}$.\\
Loss weights: $\alpha_1,\ldots,\alpha_8=\{8,7,6,5,4,3,2,1\}$.
}
\\

MP-SAE &
\parbox[t]{0.74\textwidth}{
Tied weights.\\
$T=64$ matching-pursuit steps.\\
No activation function.
}
\\

H-SAE &
\parbox[t]{0.74\textwidth}{
Upper-level TopK with $k=32$.\\
Lower-level TopK with $k=1$ per selected expert.\\
$361$ experts.\\
$16$ atoms per subspace.\\
Subspace dimension: $16$.\\
Activation function: LeakyReLU.\\
$\ell_1$ penalty: $10^{-3}$.\\
Orthogonality penalty: $10^{-1}$.
}
\\

\hline
\end{tabular}
\end{table}

\paragraph{Implementation-specific remarks.}
For TopK SAE and ArchMSAE, we did not use an auxiliary loss term to prevent dead features, which is sometimes used for these architectures. For MP-SAE, we used a fixed number of matching-pursuit steps, $T=64$, during training rather than selecting $T$ based on a reconstruction-quality threshold. For EWG-SAE, we renormalize the group sparsity weights to sum to one. This differs from the original EWG-SAE formulation, but only changes the parameterization of the sparsity penalty.

\paragraph{Training hyperparameters.}
TopK SAE, ArchMSAE, ActMSAE, and EWG-SAE follow the training approach of \citep{zaigrajew2025interpreting} (\autoref{tab:zaigrajew-training-hparams}). H-SAE was trained following \citet{muchane2025incorporating}, which use a different learning rate schedule (\autoref{tab:hsae-training-hparams}).

\begin{table}[h]
\centering
\caption{Training hyperparameters for models trained (except H-SAE)}
\label{tab:zaigrajew-training-hparams}
\begin{tabular}{ll}
\toprule
Hyperparameter & Value \\
\midrule
Batch size & $1024$ \\
Epochs & $30$ \\
Learning rate & $1\times 10^{-4}$  (except TopKSAE: $5\times 10^{-4}$)\\
\bottomrule
\end{tabular}
\end{table}

\begin{table}[h]
\centering
\caption{Training hyperparameters for H-SAE}
\label{tab:hsae-training-hparams}
\begin{tabular}{ll}
\toprule
Hyperparameter & Value \\
\midrule
Epochs & $100$ \\
Batch size & $32512$ \\
Peak learning rate & $5\times 10^{-4}$ \\
Initial learning rate & $1\times 10^{-11}$ \\
Gradient norm clipping & $0.75$ \\
Warmup steps & $200$ \\
\bottomrule
\end{tabular}
\end{table}

\subsection{Graph creation method} \label{app:graph_creation_details}
To obtain a concrete concept hierarchy graph for an SAE, we first compute feature activations over our graph-creation split of the CC3M dataset. We use these activations in the graph-creation method of \citet{wittenmayer2026insight}\footnote{
They use $
C_{ij} = \frac{\sum_{x \in D:\, A_i(x) \land A_j(x)} a_{j}(x)}{\sum_{x \in D:\, A_j(x)} a_{j}(x)}
$. $C_{ij}$ is the conditional probability of feature $i$ being active given that feature $j$ is active, weighted by the continuous activations of feature $j$ on the relevant samples.}. Since conditional probabilities involving rarely active features are unreliable, we discard features that activate on fewer than 20 of the 500{,}000 samples in the graph-creation split. As a consistent (binarized) co-activation of parents and children is essential for a well-structured hierarchy, we use a threshold of $t = 0.9$ to obtain an initial graph $G$, which is a good balance between consistent activations on the one hand and sufficiently flexible selection to obtain larger graphs on the other (cf. \autoref{fig:graph_treshold_ablation}).

Afterward, we perform several post-processing steps to denoise $G = (V, E)$ and align it with the requirements for a feature hierarchy graph. First, we prune features that only belong to small, isolated parts of the hierarchy, which are usually noisy and not semantically meaningful. 
Let $R \subseteq V$ denote the set of root nodes of the current graph, that is, the nodes with no incoming edges. For each $r \in R$, let $\mathrm{Desc}(r)$ denote the set of descendants reachable from $r$, excluding $r$ itself. We keep every root $r$ with $|\mathrm{Desc}(r)| \ge d_{\min}$ together with all nodes in $\mathrm{Desc}(r)$, and remove all remaining nodes, with $d_{\min} = 5$. We also prune all nodes with an excessively large number of children (which often correspond to noisy, always active nodes without real semantic meaning). For each node $p \in V$, we remove $p$ if $|C(p)| \geq t$, and set $t$ to 5\% of the total features in the SAE latent space.

Lastly, we compute the transitive reduction of the graph. That is, we remove every edge $(u,v) \in E$ for which there already exists a directed path of length at least $2$ from $u$ to $v$. This removes redundant edges from the graph. For example, if $p \rightarrow c$ and $c \rightarrow g$, an additional edge $p \rightarrow g$ does not change reachability but clutters the hierarchy.

\textbf{Graph creation for MP-SAEs.} When using the graph creation method for MP-SAEs, the result is normally an empty graph (or an extremely small graph when the thresholds for conditional activations are set very low). This is most likely because MP-SAEs are trained differently from other SAE approaches, particularly through the iterative reconstruction procedure. It reconstructs the input iteratively, using the remaining residual step by step, whereas other approaches reconstruct in a single pass. This can reduce co-activation of similar features (regarding the decoder column direction) and prevent hierarchical feature co-activation behavior, which is the basis for graph creation.

\subsection{Further details for Metrics} \label{app:metric_details}
A list of all requirements, their respective metrics, the graph elements they consider and the target space of the metrics are provided in \autoref{tab:desiderata_metric_overview}. For sibling similarity that uses $\mathrm{TopKInputs}_f$, we set $k=20$. This choice balances a sufficiently large input set for stable estimates while ensuring that, for all features, the selected top-$k$ inputs still correspond to sufficiently strong feature activations.

\paragraph{Post-TopK Activations}
We apply a TopK feature selection function with $k=64$ to the activations before further processing. For most of the selected SAEs, this matches the TopK sparsity constraint used during training. ActMSAE is a partial exception, as it is trained with multiple TopK levels rather than a single fixed value of $k$. Moreover, even for TopK SAE and ArchMSAE, the raw feature activations can contain positive entries beyond the $k$ features that would be retained by the TopK sparsity operator when applied with the same $k$ as during training. However, it is unclear whether these additional activations should be interpreted as meaningful. Applying TopK during evaluation therefore both aligns the activations more closely with the training setup and enforces a common sparsity level across all SAEs, improving comparability.

\paragraph{Binarization}
Both graph construction and the evaluation metrics require binarized activations. We binarize the post-TopK activations using zero as the threshold. For most inputs across all considered SAEs, this is effectively equivalent to setting the selected top-$k$ features to 1 and all remaining features to 0, as the selected features typically have positive activations.

\paragraph{Random Baseline References}
In the evaluations of our metrics, we report a random reference baseline for several values (cf.\ \autoref{fig:clip_results}). For $\texttt{s}_m$, the random reference is represented by the average similarity between two random image encodings. This is the $\texttt{s}_m$ score if the activations of all inputs were the same. For $\texttt{s}_{ha}$, the random baseline is $0.5$, which is the average if children and parents have random $\texttt{s}_m$ scores. For $\texttt{s}_{ss}$, the reference values differ because of the activation behavior of SAEs. The reference is the average feature similarity of the SAE, measured as the cosine similarity between their weighted average embeddings. As this depends on the SAE, the concrete scores vary per SAE, as shown in \autoref{tab:sibling_reference}. For $\texttt{s}_{ad}$, the reference is again $0.5$, which represents the average of a random distribution of child and parent activations.

\begin{table}[h]
\centering
\caption{Sibling similarity reference values}
\label{tab:sibling_reference}
\begin{tabular}{lcccccc}
\toprule
 & TopK SAE & EWG-SAE& ArchMSAE & ActMSAE & MP-SAE & H-SAE  \\
\midrule
CLIP & 0.6051 & 0.5812 & 0.5976 & 0.6102 & 0.6537 & 0.8468 \\
DINOv2 & 0.0255 & 0.0155 & 0.0247 & 0.0338 & 0.0299 & 0.1286 \\
Gemma-4 & 0.6358 & 0.4404 & 0.5230 & 0.4521 & 0.5860 & 0.7352 \\
\bottomrule
\end{tabular}
\end{table}

\begin{table}[]
\centering
\small
\caption{Overview of requirements, corresponding metrics, the graph elements on which the metrics operate, and their value ranges.}
\label{tab:desiderata_metric_overview}
\begin{tabular}{llll}
\toprule
\textbf{Requirement} & \textbf{Metric} & \textbf{Graph Element} & \textbf{Range} \\
\midrule
\multicolumn{4}{l}{\textit{Feature-level}} \\
Consistency & Monosemanticity Score $\texttt{s}_m$ & Concept & $[-1,1]$ \\
\midrule
\multicolumn{4}{l}{\textit{Structure-level}} \\
Hierarchical Abstractness & Hierarchical Abstractness Score $\texttt{s}_{ha}$ & Parent--child & $\{0,1\}$ \\
Sibling Similarity & Sibling Similarity Score $\texttt{s}_{ss}$ & Parent--child group & $[-1,1]$ \\
\midrule
\multicolumn{4}{l}{\textit{Co-activation-level}} \\
Activation Implication & Activation Implication Score $\texttt{s}_{ai}$ & Parent--child & $[0,1]$ \\
Activation Dominance & Activation Dominance Score $\texttt{s}_{ad}$ & Parent--child & $[0,1]$ \\
Frequent Refinement & Refinement Frequency Score $\texttt{s}_{rf}$ & Parent--child group & $[0,1]$ \\
Conditional Correlation & Conditional Correlation Score $\texttt{s}_{cc}$ & Parent--child group & $[-1,1]$ \\
\bottomrule
\end{tabular}
\end{table}

\section{Further Experimental Evaluations}

\subsection{Further Graph-level Statistics} \label{app:further_graph_level}
In the following, we provide further details on the graph statistics for all SAE methods. In addition to the graph-level statistics for CLIP reported in the main paper (\autoref{tab:clip_graph_metrics}), we provide the corresponding overview for DINOv2 in \autoref{tab:dino_graph_metrics} and for Gemma-4 E2B in \autoref{tab:gemma_graph_metrics}. The results show similar trends across all three encoders. However, the post-hoc constructed graphs for DINOv2 are generally smaller than the CLIP or Gemma-4 graphs, suggesting that the lower base similarity between DINOv2 embeddings and their differently structured encoding space reduces the number of frequently co-activating feature pairs. On two of three seeds, MP-SAE creates a non-empty graph consisting of one parent-child group. Although this is a non-empty graph for MP-SAE, it remains a very restricted, small hierarchy. 

\autoref{tab:depth_stats_clip}, \autoref{tab:depth_stats_dino} and \autoref{tab:depth_stats_gemma} report the number of nodes per hierarchy level for each method. The single node at depth 0 represents an artificially added root node that connects all other components but has no semantic significance. For all methods, the majority of nodes reside at level 1, as expected, with only TopK SAE and ActMSAE having nodes at depth $\geq 3$. The number of nodes at depth 2 for H-SAE is substantially larger than for any other method, even compared to ActMSAE, which has a similar number of parent nodes at depth 1. This suggests that while the rigid structure of H-SAE enforces consistent co-activations and refinement, it does not align well with the features the SAE naturally learns, which undermines the semantic coherence of nodes at level 2.

\begin{table}[t]
\centering
\caption{\textbf{Graph-level statistics for DINOv2.} Values are mean $\pm$ std across three seeds.  \\
\footnotesize$^\dagger$Group Size Deviation is the standard deviation of the per-seed mean group size across seeds.}
\label{tab:dino_graph_metrics}
\resizebox{\textwidth}{!}{
\begin{tabular}{@{}lrrrrrrr@{}}
\toprule
\textbf{Model}
  & \textbf{Nodes}
  & \textbf{Edges}
  & \textbf{Parent-Child}
  & \textbf{Max}
  & \textbf{Avg. Group}
  & \textbf{Group Size}
  & \textbf{Nodes at} \\
& & & \textbf{Groups} & \textbf{Depth} & \textbf{Size} & \textbf{Deviation}$^\dagger$ & \textbf{Depth $\geq 3$} \\
\midrule
TopKSAE  & $833 \pm 41$  & $1075 \pm 95$ & $124 \pm 10$ & $4.67 \pm 0.58$ & $8.64 \pm 0.16$  & $13.68 \pm 0.85$ & $102 \pm 14$ \\
ArchMSAE & $428 \pm 43$  & $546 \pm 78$  & $51 \pm \phantom{0}6$   & $2.00 \pm 0.00$ & $10.62 \pm 0.74$ & $10.78 \pm 0.54$ & $0 \pm \phantom{0}0$    \\
ActMSAE  & $1976 \pm 10$ & $3162 \pm 84$ & $586 \pm 19$ & $6.33 \pm 0.58$ & $5.39 \pm 0.06$  & $11.97 \pm 0.29$ & $315 \pm 24$ \\
H-SAE  & $6137 \pm \phantom{0}0$  & $5776 \pm \phantom{0}0$  & $361 \pm \phantom{0}0$  & $2.00 \pm 0.00$ & $16.00 \pm 0.00$ & $0.00 \pm 0.00$  & $0 \pm \phantom{0}0$    \\
EWGSAE   & $465 \pm 20$  & $1038 \pm 68$ & $91 \pm \phantom{0}6$   & $2.67 \pm 0.58$ & $11.44 \pm 0.15$ & $7.70 \pm 0.40$  & $1 \pm \phantom{0}1$    \\

\bottomrule
\end{tabular}
}
\end{table}

 \begin{table}[t]
  \centering
  \caption{\textbf{Graph-level statistics for Gemma-4 E2B.} Values are mean $\pm$ std across three seeds.  \\
  \footnotesize$^\dagger$Group Size Deviation is the standard deviation of the per-seed mean group size across seeds.}
  \label{tab:gemma_graph_metrics}
  \resizebox{\textwidth}{!}{
  \begin{tabular}{@{}lrrrrrrr@{}}
  \toprule
  \textbf{Model}
    & \textbf{Nodes}
    & \textbf{Edges}
    & \textbf{Parent-Child}
    & \textbf{Max}
    & \textbf{Avg. Group}
    & \textbf{Group Size}
    & \textbf{Nodes at} \\
  & & & \textbf{Groups} & \textbf{Depth} & \textbf{Size} & \textbf{Deviation}$^\dagger$ & \textbf{Depth $\geq 3$} \\
  \midrule
  TopK SAE & $\phantom{0}854 \pm 144$ & $1423 \pm 204$ & $\phantom{0}25 \pm \phantom{0}3$ & $3.00 \pm 0.00$ & $\phantom{0}58.06 \pm \phantom{00}8.14$ & $87.86 \pm \phantom{0}8.45$ & $\phantom{0}28 \pm \phantom{0}41$ \\
  ArchMSAE & $\phantom{0}639 \pm \phantom{0}50$ & $1389 \pm 137$ & $\phantom{0}52 \pm \phantom{0}4$ & $3.00 \pm 0.00$ & $\phantom{0}26.76 \pm \phantom{00}2.76$ & $28.38 \pm \phantom{0}5.74$ & $\phantom{0}14 \pm \phantom{00}9$ \\
  ActMSAE & $2905 \pm \phantom{0}76$ & $6347 \pm 303$ & $525 \pm 16$ & $4.67 \pm 0.58$ & $\phantom{0}12.08 \pm \phantom{00}0.25$ & $26.78 \pm \phantom{0}2.48$ & $588 \pm 128$ \\
  H-SAE & $6137 \pm \phantom{00}0$ & $5776 \pm \phantom{00}0$ & $361 \pm \phantom{0}0$ & $2.00 \pm 0.00$ & $\phantom{0}16.00 \pm \phantom{00}0.00$ & $\phantom{0}0.00 \pm \phantom{0}0.00$ & $\phantom{00}0 \pm \phantom{00}0$ \\
  MPSAE & $\phantom{0}245 \pm 216$ & $\phantom{0}244 \pm 215$ & $\phantom{00}1 \pm \phantom{0}1$ & $1.33 \pm 1.15$ & $122.00 \pm 107.53$ & $23.67 \pm 28.99$ & $\phantom{00}0 \pm \phantom{00}0$ \\
  EWG-SAE & $1096 \pm \phantom{0}69$ & $2358 \pm 173$ & $107 \pm \phantom{0}2$ & $3.00 \pm 0.00$ & $\phantom{0}21.98 \pm \phantom{00}1.73$ & $35.41 \pm \phantom{0}7.42$ & $\phantom{00}6 \pm \phantom{00}1$ \\
  \bottomrule
  \end{tabular}
  }
  \end{table}

\begin{table}[ht]
\centering
\caption{\textbf{Number of nodes at each depth level for hierarchies of SAEs trained on CLIP}. Values are mean $\pm$ std across three seeds. Missing depth levels are shown as~`--'.}
\label{tab:depth_stats_clip}
\small
\begin{tabular}{lrrrrrrrr}
\toprule
\textbf{Configuration}
  & \textbf{D0} & \textbf{D1} & \textbf{D2} & \textbf{D3}
  & \textbf{D4} & \textbf{D5} & \textbf{D6} \\
\midrule
TopK SAE  & $1 \pm 0$ & $161 \pm 11$ & $1880 \pm 51$ & $79 \pm 14$ & $1 \pm 1$   & --         & --        \\
EWG-SAE   & $1 \pm 0$ & $139 \pm 4$  & $833 \pm 51$  & $3 \pm 3$   & --          & --         & --        \\
ArchMSAE & $1 \pm 0$ & $84 \pm 4$   & $926 \pm 99$  & $2 \pm 1$   & --          & --         & --        \\
ActMSAE  & $1 \pm 0$ & $371 \pm 19$ & $1823 \pm 62$ & $949 \pm 28$& $172 \pm 19$& $22 \pm 9$ & $4 \pm 3$ \\
H-SAE  & $1 \pm 0$ & $361 \pm 0$  & $5776 \pm 0$  & --          & --          & --         & --        \\
\bottomrule
\end{tabular}
\end{table}

\begin{table}[ht]
\centering
\caption{\textbf{Number of nodes at each depth level for hierarchies of SAEs trained on DINOv2}. Values are mean $\pm$ std across three seeds.
  Missing depth levels are shown as~`--'.}
\label{tab:depth_stats_dino}
\small
\begin{tabular}{lrrrrrrrrr}
\toprule
\textbf{Configuration}
  & \textbf{D0} & \textbf{D1} & \textbf{D2} & \textbf{D3}
  & \textbf{D4} & \textbf{D5} & \textbf{D6} & \textbf{D7} \\
\midrule
TopK SAE  & $1 \pm 0$ & $78 \pm 6$   & $653 \pm 42$  & $67 \pm 15$ & $30 \pm 21$ & $6 \pm 9$  & --        & -- \\
EWG-SAE   & $1 \pm 0$ & $82 \pm 4$   & $382 \pm 17$  & $1 \pm 1$   & --          & --         & --        & -- \\
ArchMSAE & $1 \pm 0$ & $45 \pm 5$   & $383 \pm 40$  & --          & --          & --         & --        & -- \\
ActMSAE  & $1 \pm 0$ & $209 \pm 11$ & $1451 \pm 28$ & $254 \pm 11$& $44 \pm 9$  & $14 \pm 5$ & $2 \pm 1$ & $0 \pm 1$ \\
H-SAE  & $1 \pm 0$ & $361 \pm 0$  & $5776 \pm 0$  & --          & --          & --         & --        & -- \\
\bottomrule
\end{tabular}
\end{table}

\begin{table}[ht]
  \centering
  \caption{\textbf{Number of nodes at each depth level for hierarchies of SAEs trained on Gemma-4 E2B}. Values are mean $\pm$ std across three seeds. Missing depth levels are shown as~`--'.}
  \label{tab:depth_stats_gemma}
  \small
  \begin{tabular}{lrrrrrr}
  \toprule
  \textbf{Configuration}
    & \textbf{D0} & \textbf{D1} & \textbf{D2} & \textbf{D3} & \textbf{D4} & \textbf{D5} \\
  \midrule
  TopK SAE & $1 \pm 0$ & $16 \pm 3$   & $811 \pm 156$  & $28 \pm 41$   & --          & --        \\
  EWG-SAE  & $1 \pm 0$ & $82 \pm 2$   & $1008 \pm 71$  & $6 \pm 1$     & --          & --        \\
  ArchMSAE & $1 \pm 0$ & $41 \pm 2$   & $584 \pm 45$   & $14 \pm 9$    & --          & --        \\
  ActMSAE  & $1 \pm 0$ & $174 \pm 17$ & $2143 \pm 109$ & $539 \pm 120$ & $44 \pm 12$ & $4 \pm 5$ \\
  MPSAE    & $1 \pm 0$ & $1 \pm 1$ & $244 \pm 215$ & --         & --          & --        \\
  H-SAE    & $1 \pm 0$ & $361 \pm 0$  & $5776 \pm 0$   & --            & --          & --        \\
  \bottomrule
  \end{tabular}
  \\
  \end{table}

\subsection{Ablations of the Graph Creation Method}\label{app:graph_creation_ablation}

While our evaluations measure the quality of the concept hierarchy graph from all SAE methods, the results for most SAE types also depend on the underlying method to create the hierarchy graph. Next, we ablate the impact of the graph-creation threshold (\autoref{fig:graph_treshold_ablation} in the main paper), graph post-processing, and the metric used to create the graph on the insights provided in the main paper. All following results show mean and standard deviation over three seeds for SAEs evaluated on CLIP. We do not include H-SAE, as its graph is independent of post-hoc graph creation.

\begin{figure}
    \centering
    \includegraphics[width=\linewidth]{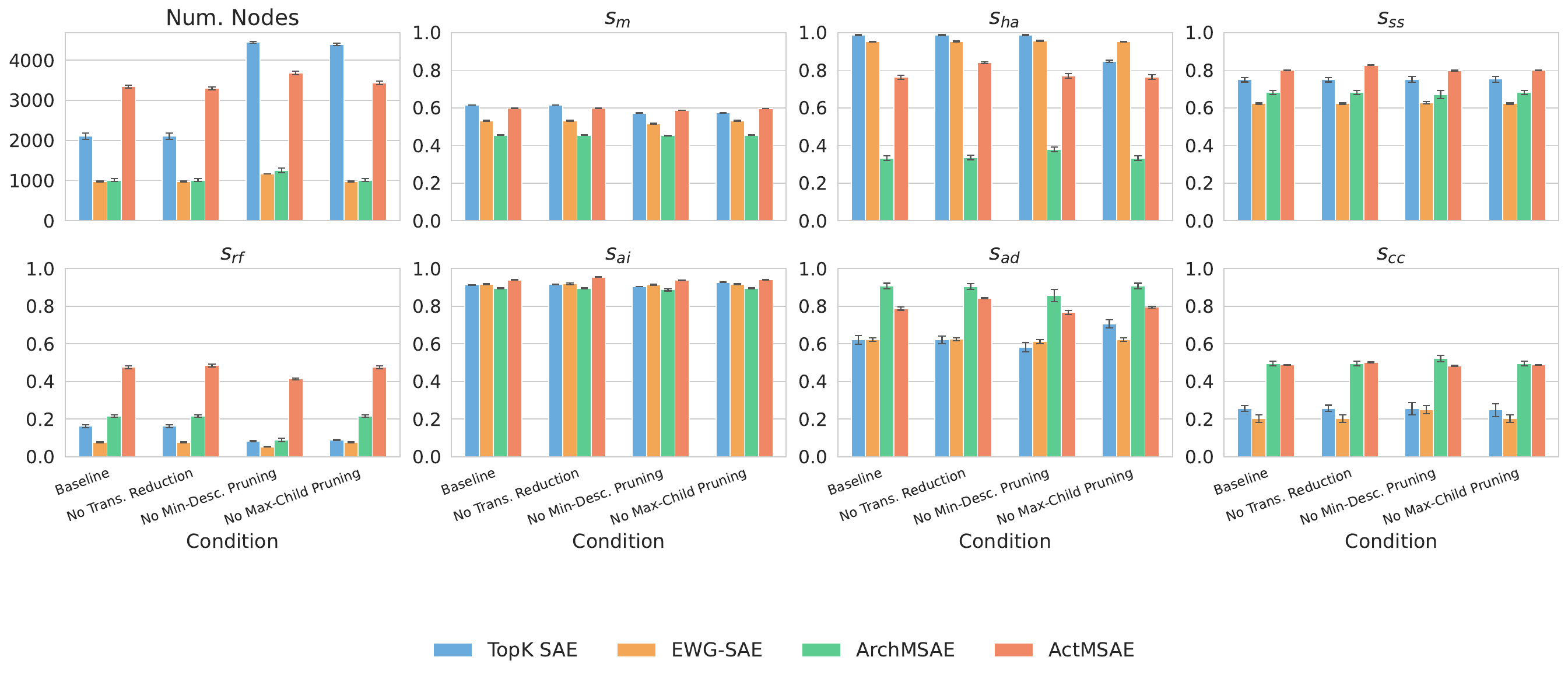}
    \caption{\textbf{Ablation of the graph post-processing on CLIP.} Mean $\pm$ standard deviation over three seeds. Post-processing steps improve hierarchy readability and quality, while trends between SAE types remain consistent.}
    \label{fig:graph_postprocessing_ablation}
\end{figure}

\paragraph{Impact of the graph post-processing.} As detailed in \autoref{app:graph_creation_details}, we introduce a post-processing step of the initial graphs to improve the quality of these graphs. Next, we show the impact of each post-processing step on hierarchy quality in \autoref{fig:graph_postprocessing_ablation}. Transitive reduction affects only ActMSAE, the only method with more than two hierarchy levels, and it leaves the hierarchy metrics essentially unchanged, so it removes less-interpretable connections at no cost to hierarchy quality. The post-processing pruning step of removing nodes with too many children mainly affects TopKSAE, and omitting it enlarges the graph but lowers $\texttt{s}_m$ and $\texttt{s}_{rf}$. Pruning small components from the hierarchies mainly affects TopKSAE and EWGSAE, and omitting it likewise enlarges the graph while lowering $\texttt{s}_m$, $\texttt{s}_{rf}$ and $\texttt{s}_{ad}$. Overall, these steps improve human-readability (transitive reduction) and hierarchy quality (the two pruning steps) without affecting the main findings.

\paragraph{Impact of the similarity metric.} The graph creation uses conditional co-activations (CCA) as metric to determine whether there is a connection between two nodes. This controls for $\texttt{s}_{ai}$ by design. In \autoref{tab:graph_metric_ablation}, we compare the quality of the resulting graphs when we use masked cosine similarity (MCS) \citep{bricken2023monosemanticity} instead of CCA.
This changes the resulting graphs but preserves the main trends seen with CCA. As expected, $\texttt{s}_{ai}$ is substantially lower under MCS, since CCA explicitly controls for this metric during graph creation; $\texttt{s}_{ad}$ and $\texttt{s}_{cc}$ are also weaker, showing that MCS-based graphs have generally worse activation behavior than CCA-based ones. Notably, MCS makes it possible to construct a graph for MPSAE (unlike CCA, whose low conditional activations prevent this). However, this MPSAE graph has near-random $\texttt{s}_{m}$ and a near-zero $\texttt{s}_{ha}$ , indicating its hierarchy lacks semantic meaning, as children are, if anything, more abstract than their parents.

 \begin{table}[t]
    \centering
    \caption{\textbf{Hierarchy metrics for graphs created with different similarity measures.} Conditional co-activations (CCA) and masked cosine similarity (MCS) on CLIP. Mean $\pm$ standard deviation over three seeds. No graph could be constructed for MPSAE with CCA because of its low conditional     activations.}
    \setlength{\tabcolsep}{3pt}
    \resizebox{\textwidth}{!}{%
    \begin{tabular}{llcccccccc}
      \toprule
      Condition & SAE & Nodes & $\texttt{s}_{m}$ & $\texttt{s}_{ha}$ & $\texttt{s}_{ss}$ & $\texttt{s}_{rf}$ & $\texttt{s}_{ai}$ & $\texttt{s}_{ad}$ & $\texttt{s}_{cc}$ \\
      \midrule
      CCA & TopK SAE  & 2120\,{\scriptsize$\pm$\,42} & 0.616\,{\scriptsize$\pm$\,0.001} & 0.986\,{\scriptsize$\pm$\,0.003} & 0.750\,{\scriptsize$\pm$\,0.012} & 0.162\,{\scriptsize$\pm$\,0.008} & 0.914\,{\scriptsize$\pm$\,0.002} &
  0.621\,{\scriptsize$\pm$\,0.023} & 0.257\,{\scriptsize$\pm$\,0.017} \\
      MCS & TopK SAE  & 1567\,{\scriptsize$\pm$\,93} & 0.564\,{\scriptsize$\pm$\,0.001} & 0.858\,{\scriptsize$\pm$\,0.024} & 0.766\,{\scriptsize$\pm$\,0.004} & 0.385\,{\scriptsize$\pm$\,0.011} & 0.587\,{\scriptsize$\pm$\,0.015} &
  0.738\,{\scriptsize$\pm$\,0.004} & 0.301\,{\scriptsize$\pm$\,0.009} \\
      \midrule
      CCA & ArchMSAE & 1012\,{\scriptsize$\pm$\,99} & 0.455\,{\scriptsize$\pm$\,0.000} & 0.332\,{\scriptsize$\pm$\,0.012} & 0.682\,{\scriptsize$\pm$\,0.012} & 0.215\,{\scriptsize$\pm$\,0.007} & 0.895\,{\scriptsize$\pm$\,0.002} &
  0.907\,{\scriptsize$\pm$\,0.015} & 0.495\,{\scriptsize$\pm$\,0.013} \\
      MCS & ArchMSAE & 2517\,{\scriptsize$\pm$\,15} & 0.441\,{\scriptsize$\pm$\,0.000} & 0.521\,{\scriptsize$\pm$\,0.017} & 0.700\,{\scriptsize$\pm$\,0.005} & 0.352\,{\scriptsize$\pm$\,0.008} & 0.324\,{\scriptsize$\pm$\,0.002} &
  0.667\,{\scriptsize$\pm$\,0.003} & 0.369\,{\scriptsize$\pm$\,0.008} \\
      \midrule
      CCA & ActMSAE  & 3341\,{\scriptsize$\pm$\,19} & 0.598\,{\scriptsize$\pm$\,0.001} & 0.763\,{\scriptsize$\pm$\,0.011} & 0.800\,{\scriptsize$\pm$\,0.001} & 0.476\,{\scriptsize$\pm$\,0.007} & 0.940\,{\scriptsize$\pm$\,0.001} &
  0.788\,{\scriptsize$\pm$\,0.009} & 0.487\,{\scriptsize$\pm$\,0.003} \\
      MCS & ActMSAE  & 4459\,{\scriptsize$\pm$\,19} & 0.567\,{\scriptsize$\pm$\,0.001} & 0.853\,{\scriptsize$\pm$\,0.001} & 0.722\,{\scriptsize$\pm$\,0.001} & 0.480\,{\scriptsize$\pm$\,0.004} & 0.154\,{\scriptsize$\pm$\,0.002} &
  0.683\,{\scriptsize$\pm$\,0.008} & 0.368\,{\scriptsize$\pm$\,0.003} \\
      \midrule
      CCA & EWG-SAE   & \phantom{0}974\,{\scriptsize$\pm$\,57} & 0.530\,{\scriptsize$\pm$\,0.003} & 0.953\,{\scriptsize$\pm$\,0.002} & 0.624\,{\scriptsize$\pm$\,0.004} & 0.075\,{\scriptsize$\pm$\,0.003} & 0.918\,{\scriptsize$\pm$\,0.004} &
  0.622\,{\scriptsize$\pm$\,0.009} & 0.202\,{\scriptsize$\pm$\,0.020} \\
      MCS & EWG-SAE   & 1368\,{\scriptsize$\pm$\,25} & 0.501\,{\scriptsize$\pm$\,0.003} & 0.681\,{\scriptsize$\pm$\,0.030} & 0.636\,{\scriptsize$\pm$\,0.008} & 0.258\,{\scriptsize$\pm$\,0.012} & 0.391\,{\scriptsize$\pm$\,0.012} &
  0.531\,{\scriptsize$\pm$\,0.014} & 0.223\,{\scriptsize$\pm$\,0.008} \\
      \midrule
      CCA & MPSAE    & --- & --- & --- & --- & --- & --- & --- & --- \\
      MCS & MPSAE    & \phantom{0}504\,{\scriptsize$\pm$\,23} & 0.472\,{\scriptsize$\pm$\,0.003} & 0.030\,{\scriptsize$\pm$\,0.008} & 0.797\,{\scriptsize$\pm$\,0.013} & 0.398\,{\scriptsize$\pm$\,0.009} & 0.122\,{\scriptsize$\pm$\,0.003} &
  0.942\,{\scriptsize$\pm$\,0.009} & 0.336\,{\scriptsize$\pm$\,0.008} \\
      \bottomrule
    \end{tabular}}
    \label{tab:graph_metric_ablation}
  \end{table}

\subsection{Validating Hierarchy Requirements with a Ground-Truth Hierarchy.}
The requirements and metrics introduced in this work are intentionally designed to be evaluated unsupervised, without the need for detailed human hierarchy labels. While this makes them more applicable in many practical scenarios where ground-truth hierarchies are not known, it is still possible to validate the structure-level metrics against known ground-truth hierarchies. To do so, we use the well-known ImageNet-1k class hierarchy \citep{deng2009imagenet} and evaluate our metrics on images from ImageNet directly. We randomly sample up to 1000 images per class and use class membership as a binary neuron activation for the metric computations. For $\texttt{s}_{ss}$, we use the average embedding of 20 randomly selected images per class as the "top-activating" images.

The results in \autoref{tab:imagenet_hierarchy_eval} show a strong connection between the monosemanticity score and hierarchy level, indicated by an $\texttt{s}_{ha}$ of nearly 1. This supports our claim that specificity (as measured by $\texttt{s}_m$) increases through the hierarchy, and confirms that $\texttt{s}_{ha}$ is a reasonable measure in this context. Further, the similarity between siblings in this hierarchy is captured by $\texttt{s}_{ss}$.

\begin{table}[]
    \centering
    \caption{\textbf{Evaluating the ImageNet-1k hierarchy.} Applying our concept- and structure-level metrics to the ImageNet-1k hierarchy indicates a strong connection between the desired metrics and the ground-truth hierarchy structure.}
    \begin{tabular}{l|rrr}
    \toprule
    Metric  & $\texttt{s}_m$ & $\texttt{s}_{ha}$ & $\texttt{s}_{ss}$ \\
    Mean     &  $0.6855$ & $0.9401$ & $0.8639$ \\
    \bottomrule
    \end{tabular}
    \label{tab:imagenet_hierarchy_eval}
\end{table}

\subsection{Investigating the Connection between Sibling Similarity and Feature Redundancy}\label{sec:sibling_redundancy}

Sibling similarity itself does not check for near-duplicate features, since a feature is trivially most similar to itself. To further quantify how this metric relates to redundancy, we additionally report the Jaccard index \citep{costa2021further} as a redundancy measure between siblings (cf. \autoref{tab:sibling-jaccard}, left). This measures how often two siblings co-activate relative to how often at least one activates, and has also been used in the context of SAEs \citep{pach2025sparse}.
For around 1700 features of ActMSAE on CLIP and around 700 features on DINO, the Jaccard similarity to the closest sibling exceeds 0.5 and is potentially redundant. This occurs primarily at lower levels of the hierarchy, where increasingly similar features get grouped together (cf. \autoref{tab:sibling-jaccard}, right).

Overall, while most SAE features remain quite distinct, features at lower depths become increasingly similar, making it harder to automatically distinguish highly similar from truly redundant features, which is an important general question for SAE feature spaces rather than a weakness specific to our approach.

\begin{table}[t]
\centering
\caption{Sibling Jaccard similarity. \textbf{Left:} Average Jaccard similarity between siblings for each SAE type on CLIP ViT-L/14 and DINOv2-base. \textbf{Right:} Average Jaccard similarity to the closest sibling of ActMSAE features by hierarchy depth. (Values at depth $\geq 5$ are unreliable due to low feature counts at that depth.)}
\begin{minipage}[t]{0.48\linewidth}
  \centering
  \begin{tabular}{lcc}
    \toprule
    SAE type & CLIP ViT-L/14 & DINOv2-base \\
    \midrule
    TopK SAE & 0.024 & 0.095 \\
    EWG-SAE  & 0.008 & 0.012 \\
    ActMSAE  & 0.320 & 0.214 \\
    ArchMSAE & 0.023 & 0.071 \\
    H-SAE     & 0.000 & 0.021 \\
    \bottomrule
  \end{tabular}
  \label{tab:sibling-jaccard-sae}
\end{minipage}\hfill
\begin{minipage}[t]{0.48\linewidth}
  \centering
  \begin{tabular}{lcc}
    \toprule
    Depth & CLIP ViT-L/14 & DINOv2-base \\
    \midrule
    1 & 0.065 & 0.148 \\
    2 & 0.485 & 0.271 \\
    3 & 0.476 & 0.361 \\
    4 & 0.457 & 0.434 \\
    5 & 0.457 & 0.246 \\
    6 & 0.459 & 0.522 \\
    \bottomrule
  \end{tabular}
  \label{tab:sibling-jaccard-depth}
\end{minipage}
\label{tab:sibling-jaccard}
\end{table}

\subsection{Connection to Hierarchical Clustering Metrics}
While metrics for hierarchical clustering are related to concept hierarchies, there are structural differences in the assumptions of the hierarchy elements.
The basic elements of our feature hierarchy graph are features, compared to samples in a dendrogram. Intermediate nodes are also meaningful higher-level features, whereas dendrogram intermediate nodes are purely structural connection points. Further, our hierarchy graph is not a strict tree but a DAG with potentially disconnected components, where edges carry explicit semantic meaning rather than representing similarity-based merges.

These differences make a direct adaptation of hierarchical clustering metrics difficult and require a small adaptation of the cophenetic correlation to apply it to feature hierarchy graphs.

The original cophenetic correlation coefficient between a dissimilarity matrix $x$ and the cophenetic distances $t$ of a dendrogram is
\begin{equation}
  c = \frac{\sum_{i<j} \bigl(x(i,j) - \bar{x}\bigr)\bigl(t(i,j) - \bar{t}\bigr)}
           {\sqrt{\sum_{i<j} \bigl(x(i,j) - \bar{x}\bigr)^2 \,\sum_{i<j} \bigl(t(i,j) - \bar{t}\bigr)^2}},
  \label{eq:cophenetic}
\end{equation}
where $x(i,j)$ is the original dissimilarity between leaves $i$ and $j$, $t(i,j)$ is the cophenetic distance (the height of the dendrogram at which $i$ and $j$ are first merged), and $\bar{x}$, $\bar{t}$ are the means of $x(i,j)$ and $t(i,j)$ over all pairs $i<j$. In our adaptation, $t(i,j)$ is replaced by the depth of $\mathrm{LCA}(i,j)$ and $x(i,j)$ by the pairwise similarity like in $\texttt{s}_{ss}$. Since all SAEs except ActMSAE have only two levels, this reduces to measuring whether more similar nodes share a parent or fall under different parents (where the LCA is the artificial root at depth $0$), making it closely related to $\texttt{s}_{\mathrm{ss}}$.

\begin{table}[t]
  \centering
  \caption{Number of leaves and cophenetic correlation for each SAE type on CLIP.}
  \begin{tabular}{lccccc}
    \toprule
     & TopK SAE & EWG-SAE & ArchMSAE & ActMSAE & H-SAE \\
    \midrule
    Leaves                 & 1884  & 808   & 922   & 1971  & 5776  \\
    Cophenetic correlation & 0.187 & 0.022 & 0.220 & 0.252 & 0.077 \\
    \bottomrule
  \end{tabular}
  \label{tab:cophenetic}
\end{table}

The results in \autoref{tab:cophenetic} then also match our evaluation on CLIP: EWGSAE's and H-SAE's sibling similarities are close to random, while ActMSAE's is highest, and highlighting the connection between both metrics.

\subsection{More Qualitative Illustrations}\label{app:more_qualitative}

For the illustrative examples, we show a representative subset of child features and not necessarily all examples to their lowest depth to maintain readability.

\begin{figure}
    \centering
    \begin{subfigure}[t]{0.6\linewidth}
        \centering
        \includegraphics[height=3.5cm]{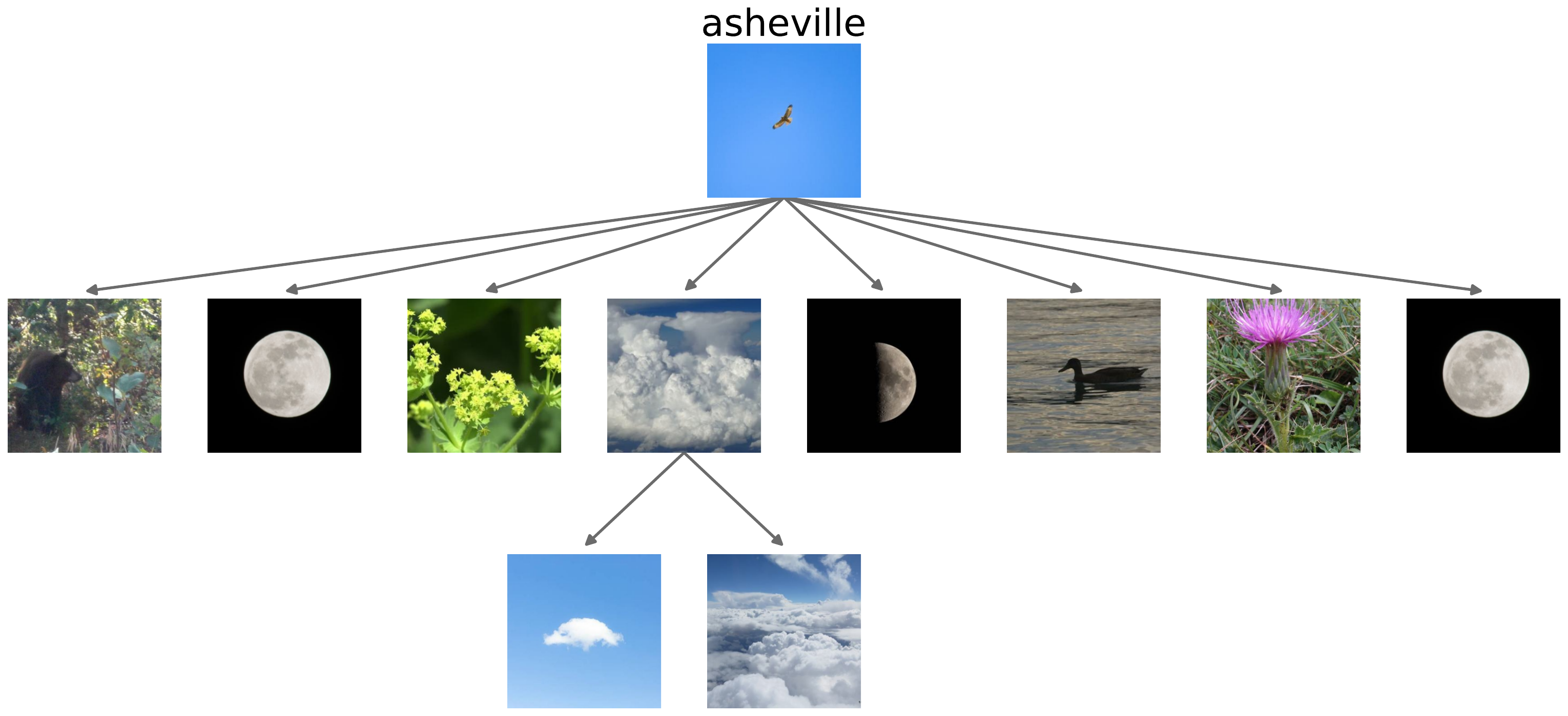}
    \end{subfigure}
    \hfill
    \begin{subfigure}[t]{0.35\linewidth}
        \centering
        \includegraphics[height=3.5cm]{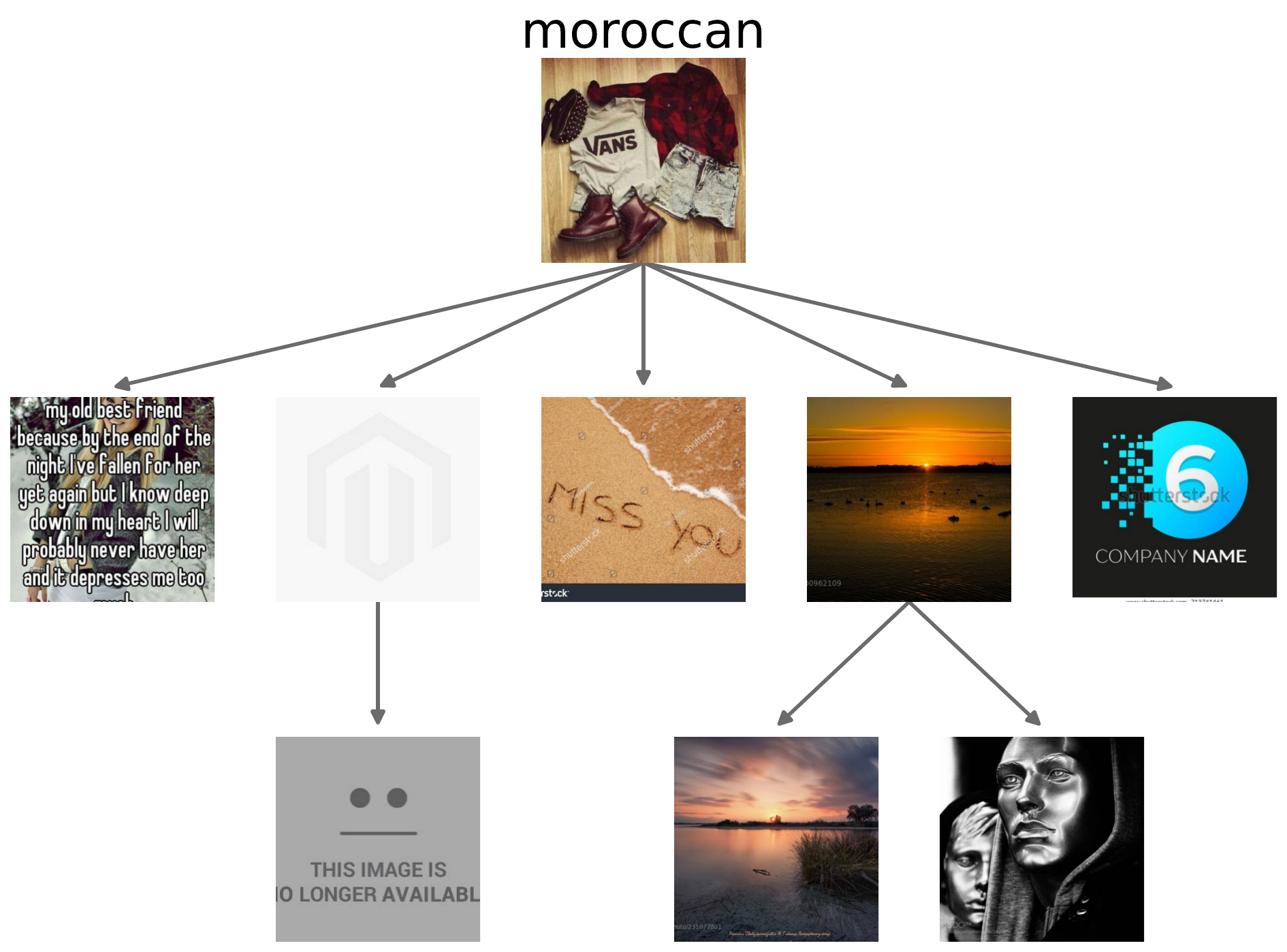}
    \end{subfigure}
    \caption{\textbf{Failure modes of learned hierarchies.} Left: children are grouped under a common parent based on spurious co-occurrences rather than clear semantic connections. Right: several child features are themselves noisy, for example, activating on placeholder sprites, which could be mitigated by feature-interpretability filtering prior to graph construction. All examples are taken from ActMSAE trained on CLIP.}
    \label{fig:qualitative_neg_actmsae}
\end{figure}

\paragraph{Semantic failure modes of learned hierarchies.}
To complement the qualitative examples in the main paper (cf.\ \autoref{sec:experiments}), we provide additional examples here. While most ActMSAE hierarchy subgraphs are human-interpretable and semantically sensible, there are also cases where the learned hierarchies are harder to follow or lack clear meaning. \autoref{fig:qualitative_neg_actmsae} shows two such failure cases. In the left example, the automated naming strategy of \citet{rao2024discover} assigns the label \textit{asheville} to a concept primarily centered around birds in flight against open sky. The children have only a loose semantic connection to the parent and low mutual similarity, covering concepts such as moon imagery, plants, and birds. The right example groups several concepts under \textit{moroccan}, where most children are themselves noisy, consisting, for instance, of placeholder sprites. These two examples illustrate distinct failure modes: the first groups seemingly unrelated concepts based on spurious co-occurrences, while the second contains low-quality features that could have been filtered out by feature-interpretability evaluations prior to graph construction.

\begin{figure}
    \centering
    \begin{subfigure}[c]{0.55\linewidth}
        \centering
        \includegraphics[height=2.2cm]{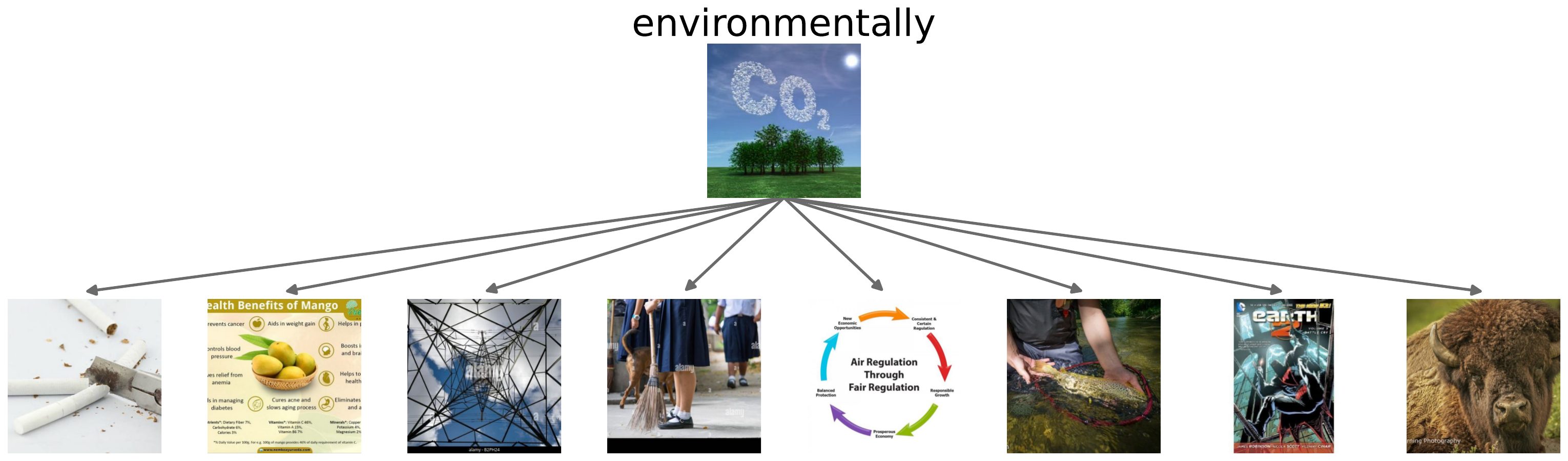}
        \caption{H-SAE Subgraph Example}
    \end{subfigure} \hfill
    \begin{subfigure}[c]{0.2\linewidth}
        \centering
        \includegraphics[height=2.2cm]{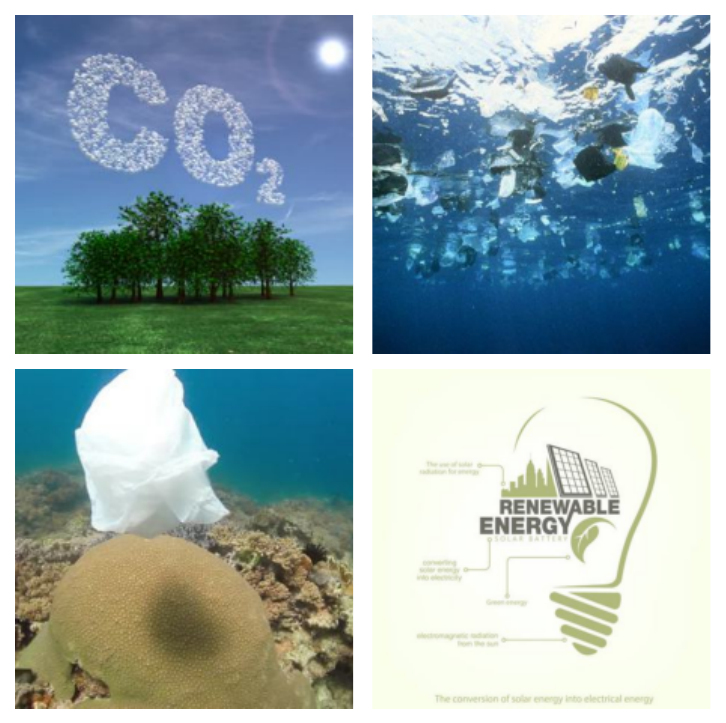}
        \caption{Parent}
    \end{subfigure}
        \begin{subfigure}[c]{0.2\linewidth}
        \centering
        \includegraphics[height=2.2cm]{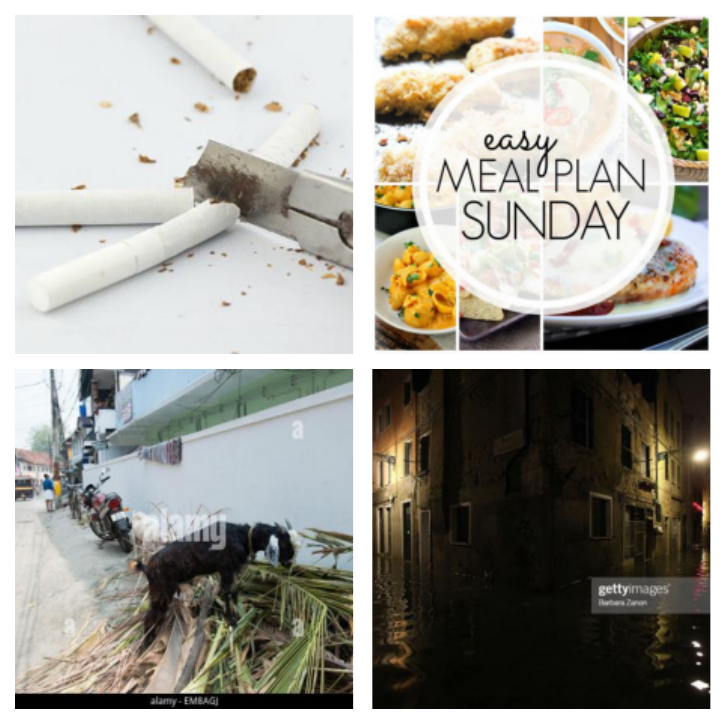}
        \caption{Leftmost child}
    \end{subfigure}
    \caption{\textbf{Qualitative example of H-SAE's failure mode.} The parent concept represents a coherent concept about the environment and its pollution, while the children do not have clear semantic connections to the parent. Even more, the top-activating images of the leftmost child themselves do not show any coherent concept.}
    \label{fig:qualitative_hsae_1}
\end{figure}

\begin{figure}
    \centering
    \begin{subfigure}[c]{0.55\linewidth}
        \centering
        \includegraphics[height=2.2cm]{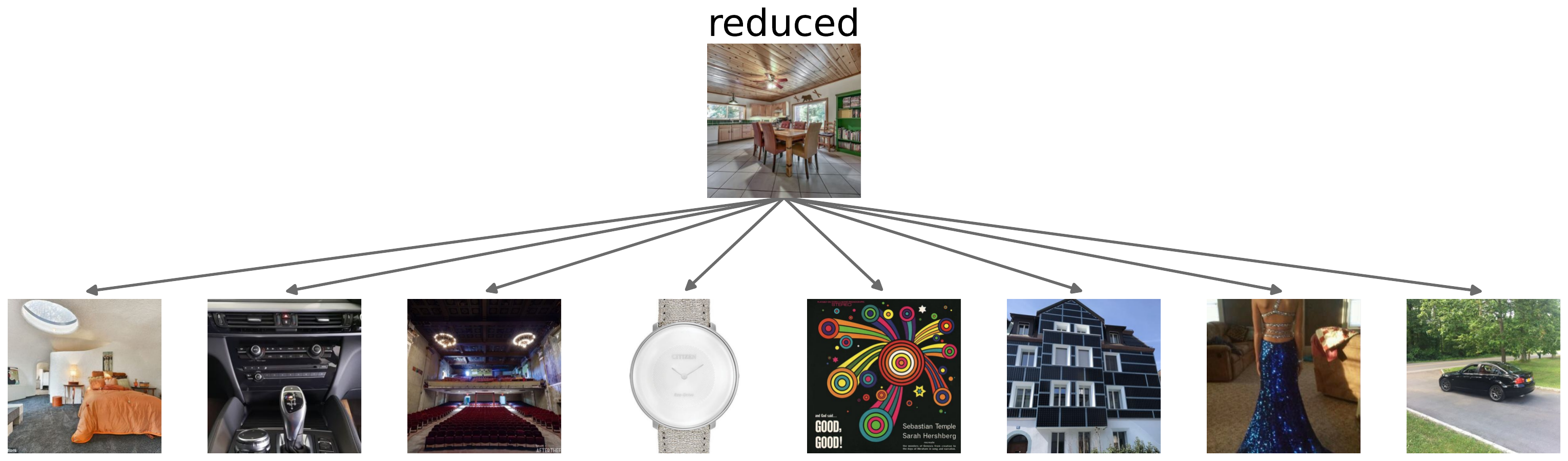}
        \caption{H-SAE Subgraph Example}
    \end{subfigure} \hfill
    \begin{subfigure}[c]{0.2\linewidth}
        \centering
        \includegraphics[height=2.2cm]{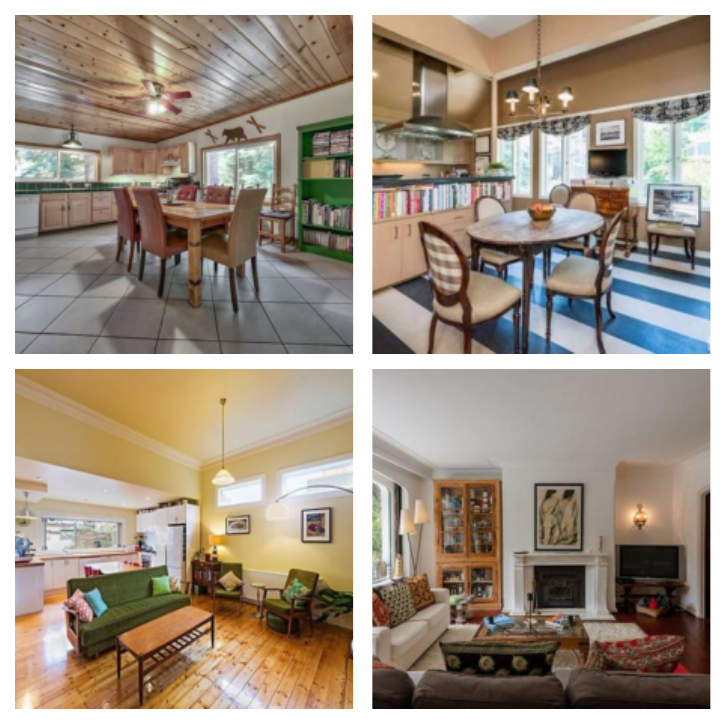}
        \caption{Parent}
    \end{subfigure}
        \begin{subfigure}[c]{0.2\linewidth}
        \centering
        \includegraphics[height=2.2cm]{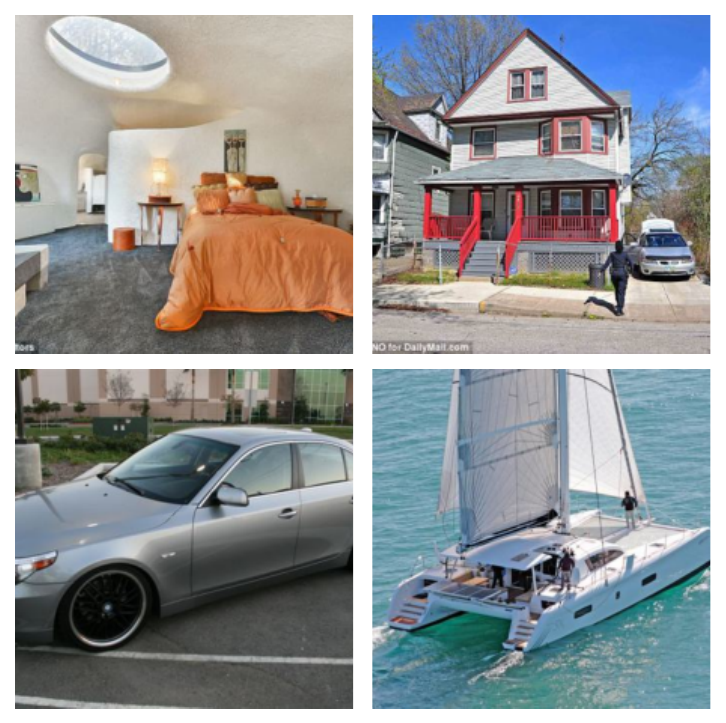}
        \caption{Leftmost child}
    \end{subfigure}
    \caption{\textbf{Further qualitative example of H-SAE.} In this example, the parent concept represents a living room or a kitchen, while the children again do not have clear semantic connections to the parent. Similarly, the top-activating images of the leftmost child do not show a coherent concept.}
    \label{fig:qualitative_hsae_2}
\end{figure}

\paragraph{Illustration of H-SAE's Failure Mode.}
While H-SAE produces a fixed graph structure with consistent parent-child co-activations, our evaluation highlights a systematic lack of hierarchical semantics. Specifically, the hierarchical abstractness score $\texttt{s}_{ha}$ reveals that parent features are typically more specific than their children — the opposite of what a generalization/specialization hierarchy requires. The underlying cause is that child features often fail to represent coherent semantic concepts, resulting in near-random $\texttt{s}_m$ scores and consequently near-zero $\texttt{s}_{ha}$. This is, therefore, less a failure of hierarchical structure per se than a general quality problem with the learned child features.
\autoref{fig:qualitative_hsae_1} and \autoref{fig:qualitative_hsae_2} illustrate this: the parent concepts are somewhat coherent, environmental scenes and pollution in the first example, indoor kitchens and living rooms in the second, but the children bear no clear semantic connection to their parents. The top-activating images for the child features themselves show no discernible underlying concept, underscoring that child-feature interpretability is the core limitation of this approach.

\begin{figure}
    \centering
    \includegraphics[width=0.9\linewidth]{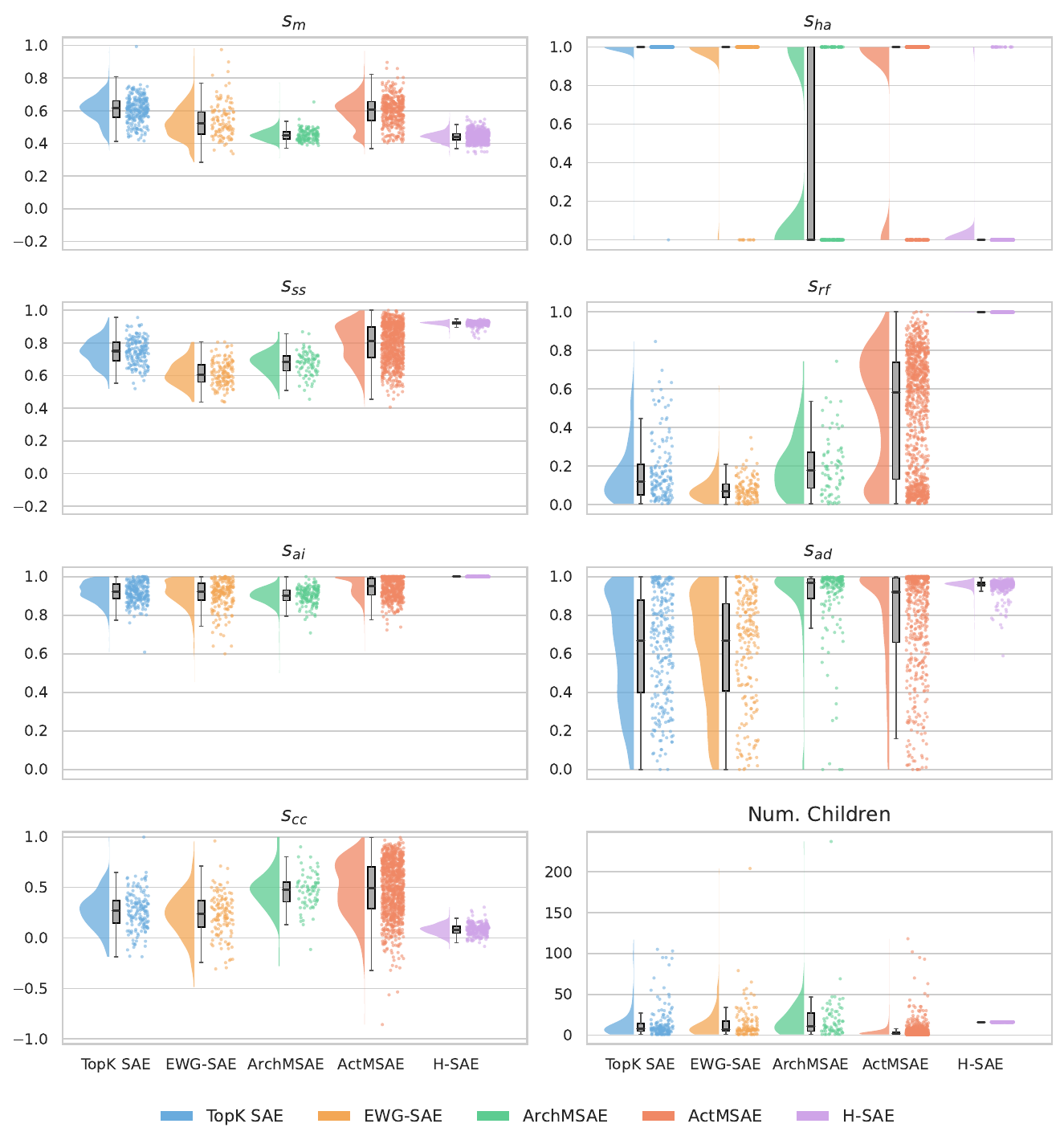}
    \caption{\textbf{Distribution of the metrics per graph element for SAEs trained on CLIP.} While the distributions of nodes, edges, or parent-child groups are mostly well-behaved, there is a clear two-modal distribution for ActMSAE visible, most likely due to the big differences between hierarchy depths. ($\texttt{s}_{ha}$ takes only discrete values, meaning that the raincloud plots for this metric do not provide more information than the mean). 
    }
    \label{fig:raincloud_clip}
\end{figure}

\begin{figure}
    \centering
    \includegraphics[width=0.9\linewidth]{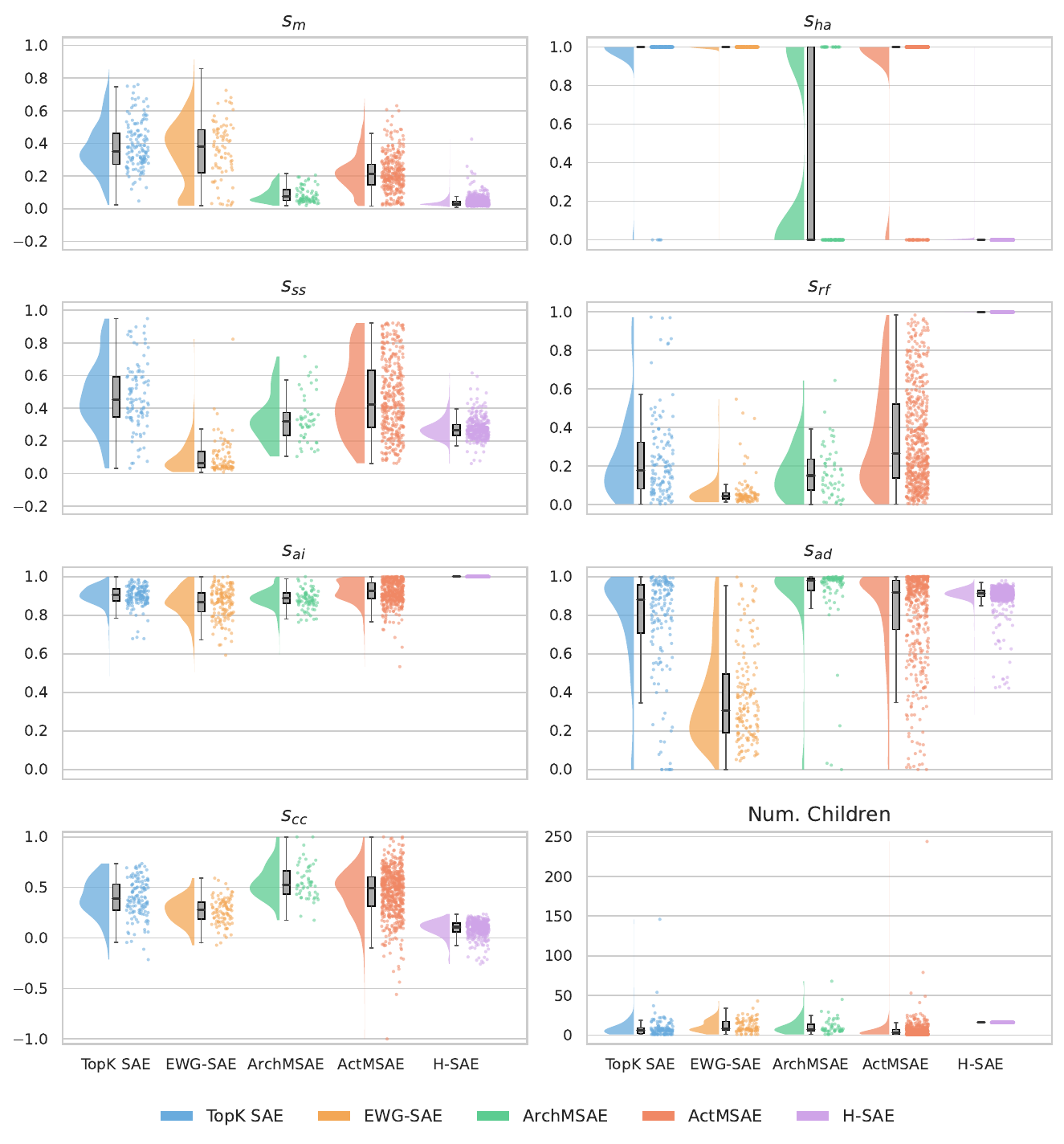}
    \caption{\textbf{Distribution of the metrics per graph element for SAEs trained on DINOv2.}}
    \label{fig:raincloud_dino}
\end{figure}

\paragraph{Analyzing the distributions of the results.}
In the main evaluation, hierarchy metrics are reported as means and standard deviations over a single graph. Here, we examine the distribution of individual scores within each graph in more detail. \autoref{fig:raincloud_clip} and \autoref{fig:raincloud_dino} show the results for CLIP and DINOv2 respectively. For most metrics and methods, the distributions are approximately unimodal and well-behaved. Two notable exceptions stand out, however. First, ActMSAE shows bimodal behavior across several metrics, likely reflecting the difference between hierarchy levels observed in \autoref{fig:depth_analysis}: top-level graph elements tend to have lower $\texttt{s}_m$ and lower $\texttt{s}_{rf}$, pulling the distribution apart into two groups. Second, while the mean $\texttt{s}_{cc}$ is above zero for all SAEs, each method has a substantial number of parent-child groups with negative conditional correlation, which means that parent activations decrease as child activations grow. This is precisely the soft absorption behavior our metrics are designed to capture, and its prevalence across methods underscores that it remains an unsolved challenge for current hierarchical SAE approaches.

\end{document}